\pdfoutput=1
\documentclass[11pt]{article}

\usepackage{acl}

\usepackage{natbib}
\setcitestyle{sort}
\usepackage{times}
\usepackage{latexsym}
\usepackage[T1]{fontenc}
\usepackage[utf8]{inputenc}

\usepackage{microtype}

\usepackage{epigraph}
\usepackage{dirtytalk}
\usepackage{csquotes}
\usepackage{enumitem}

\title{What company do words keep? \\ Revisiting the distributional semantics of J.R. Firth \& Zellig Harris}

\author{Mikael Brunila \\
  McGill University\\
  Department of Geography\\
  \texttt{mikael.brunila@gmail.com} \\\And
  Jack LaViolette \\
  Columbia University \\
  Department of Sociology \\
  \texttt{jack.laviolette@columbia.edu} \\}

\begin{document}
\maketitle
\begin{abstract}
The power of word embeddings is attributed to the linguistic theory that similar words will appear in similar contexts. This idea is specifically invoked by noting that ``you shall know a word by the company it keeps,'' a quote from British linguist J.R. Firth who, along with his American colleague Zellig Harris, is often credited with the invention of ``distributional semantics.'' While both Firth and Harris are cited in all major NLP textbooks and many foundational papers, the content and differences between their theories is seldom discussed. Engaging in a close reading of their work, we discover two distinct and in many ways divergent theories of meaning. One focuses exclusively on the internal workings of linguistic forms, while the other invites us to consider words in new company---not just with other linguistic elements, but also in a broader cultural and situational context. Contrasting these theories from the perspective of current debates in NLP, we discover in Firth a figure who could guide the field towards a more culturally grounded notion of semantics. We consider how an expanded notion of ``context'' might be modeled in practice through two different strategies: comparative stratification and syntagmatic extension.
\end{abstract}

\section{Introduction}

\epigraph{We are in the world and the world is in us.}{Alfred North Whitehead \\ (\citeyear{whitehead_modes_1938}; cited in \citealt[29]{firth_synopsis_1957})}

If you have read any papers in computational linguistics in the past thirty years, you have likely come upon the following quote from British linguist J.R.Firth (\citeyear[11]{firth_synopsis_1957}): ``You shall know a word by the company it keeps''. Cited in most major textbooks \cite{manning_foundations_1999,jurafsky_speech_2009,eisenstein_introduction_2019,russell_artificial_2020}, several foundational papers, and hundreds of other NLP articles, this phrase has come to index a theoretical orientation in a field that is increasingly focused on computation, often at the expense of linguistic theory  \cite[on these trends in NLP see][]{halevy_unreasonable_2009,manning_computational_2015,norvig_colorless_2012,henderson_unstoppable_2020,church2021future}. Together with American linguist Zellig Harris, Firth is regularly called upon to justify a distributional theory of semantics, whereby the meaning of lexical units is conceived in terms of relative co-occurrence and shared contexts of use \cite{sahlgren_distributional_2008}.

While Harris and Firth are often invoked, their ideas are seldom closely engaged. Hailing from disparate traditions, Harris and Firth had radically different ideas on the scope and context of linguistic analysis, and presented incongruent versions of the distributional method. Drawing on the information theory pioneered by Claude Shannon (\citeyear{shannon_mathematical_1948}), Harris was determined to work out a structuralist theory of language in terms of mathematical information \cite{leon_z_2011, nevin_minimalist_1993}. Firth, on the other hand, came to linguistics via anthropology and borrowed heavily from pragmatic philosophies of language. For him, linguistic analysis always started with the ``context of situation'' and necessarily accounted for non-verbal actors and objects \citep[][9]{firth_synopsis_1957}.

Considering the definition and extent of linguistic context is important for many reasons, as a spate of recent publications suggests \citep{glenberg00,hovy_social_2018,bender_climbing_2020,bisk_experience_2020,   tamari_language_2020,trott_reconstruing_2020}. Firstly, it touches upon the limits of current paradigms in NLP, where corpus linguistics is perfected through increasingly complex models trained on increasingly massive corpora \cite{bender_dangers_2021}. This approach may advance the identification of linguistic \textit{form}, but might ultimately have little to say about the relation of \textit{meaning} to the social world \cite{bender_climbing_2020, bisk_experience_2020}. Secondly, even with more modest ambitions, several NLP applications---e.g., with spatial \cite{mckenzie_natural_2021} or historical \cite{kutuzov_diachronic_2018} data---require that linguistic patterns be related to other types of structure. Thirdly, from sociological and sociolinguistic perspectives, meaning intrinsically varies as language is used in different settings and indexed to different social categories \cite{labov_sociolinguistic_1972,bourdieu_distinction_1984, silverstein_indexical_2003, eckert_variation_2008,  hovy_social_2018}. Finally, without a broader sense of context, NLP and language modeling in particular remains trapped in a paradigm where language is always treated as universal, making invisible both different speech communities  \cite[e.g.][]{nguyen_learning_2021} and the biases of language \cite[e.g.][]{ blodgett_language_2020,lu_gender_2020, sap_social_2020}.

After a brief history of distributional semantics, we outline Harris' and Firth's research on distribution and, more broadly, on the scope of linguistic analysis. We look at some of the ways in which NLP has tried to account for broader context within the distributional paradigm. We suggest that existing strategies can be understood in terms of either ``comparative stratification'' or ``syntagmatic extension.'' We conclude with thoughts on why re-reading Harris and, in particular, Firth might aid the field of NLP with its current aporias. If words shall be known by the company they keep, then the question follows: what \textit{kind} of company do they keep? Are they found only alongside linguistic elements, or do they mingle with other types of entities? Or, as Firth himself wrote: ``Many different answers could be given to the question `Distribution of what, where and how?''' \cite[$v$]{firth_introduction_1957}.

\section{Background: Distributional Semantics and NLP}

Distributional semantics has been an fundamental part of computational linguistics since the beginnings of the field, but in a discontinuous manner encompassing at least two distinct eras. Firstly, during the 1950s and 60s, Harris was integral to the mathematization of linguistics in the US after the Second World War \cite{rubenstein_contextual_1965, leon_automating_2021}. Firth was skeptical of efforts to mechanize linguistics,\footnote{He seemed to consider the idea Orwellian \cite{firth_modes_1957} and repeatedly attacked Norbert Wiener \cite[e.g.][]{firth_descriptive_1968, firth_linguistic_1968}, a pioneer whose work would later be considered foundational for connectionist AI \cite{goodfellow2016deep, russell_artificial_2020}.} but he nonetheless consulted for some of the early work on machine translation at Cambridge Language Research Group \cite[410]{leon_linguistic_2007}, which included Firth's pupil, M.A.K. Halliday \cite[144]{leon_automating_2021} and shortly later the NLP pioneer Karen Sp\"{a}rck Jones \cite[89]{leon_automating_2021}. Naturally, others also contributed to this first wave of distributional thinking, including Shannon (\citeyear{shannon_mathematical_1945,shannon_mathematical_1948}) with what might be considered one of the first language models, Warren Weaver (\citeyear{weaver_translation_1952}) with an early proposal for distributional semantics, and Martin Joos (\citeyear{joos_description_1950}) with a statistical formulation of language as a symbolic system of conditional probabilities. 

Secondly, when computational linguistics returned to its ``empiricist'' roots in probabilistic methods and information theory in the mid-80s and early 90s (\citealt{norvig_colorless_2012}; \citealt[141]{leon_automating_2021}), Firth and Harris accompanied Shannon among the authors who were invoked, in an ACL ``Special Issue on Computational Linguistics Using Large Corpora,'' as foundational figures of a tradition that had been overshadowed for decades by the ``rationalism'' of characters such as Noam Chomsky and Marvin Minsky \cite[15]{church_introduction_1993}. During this ``corpus turn,'' the rapid automation of linguistics was driven by a resumed connection with postwar computational linguistics and information theory \cite[3]{leon_automating_2021}. However, the 1990s wave of vector semantics papers that used methods like singular-value decomposition (SVD) to produce early ``dense vector'' models of meaning like LSA \cite{deerwester_computer_1989,deerwester_indexing_1990,landauer_solution_1997} and its derivatives \cite{hofmann_probabilistic_1999,blei_latent_2002}, HAL \cite{burgess1998simple}, or the models of Schütze \cite{schutze_dimensions_1992,schutze_word_1993,schutze_vector_1993} generally did not cite Firth or Harris, although a few papers from that period did \cite{church_word_1989,Hindle90}. In short, while Firth and Harris were not regularly used as stand-ins for linguistic theory during the 1990s and early 2000s, a general revival of empiricism and distributional approaches to meaning signaled a potential resurgence of interest in their thinking.

During the 2000s, the application of neural networks to language modeling tasks \cite[e.g.][]{bengio_neural_2003} and the development of self-supervision techniques \cite[e.g.][]{raina_self-taught_2007} set the stage for the word embedding breakthroughs of the early 2010s \cite[e.g.][]{mikolov_distributed_2013}. By the end of the decade, the introduction of attention \cite{graves_speech_2013, bahdanau2015neural} and then of the Transformer model \cite{vaswani_attention_2017} made way for the next breakthrough, the large-language modeling revolution \cite[e.g.][]{peters_deep_2018,devlin_bert_2019,chowdhery_palm_2022}.

Following the introduction of the word2vec model and its powerful but ``static'' embeddings, Harris in particular was frequently cited \cite{le_distributed_2014, levy_dependency-based_2014, levy_improving_2015,bojanowski_enriching_2017}, often (but not always) along with Firth \cite{bruni_multimodal_2014,hamilton_diachronic_2016,goldberg_neural_2017,  eisenstein_introduction_2019,jurafsky_speech_2021}. However, despite an explosion of citations \cite[8719]{bisk_experience_2020}, this interest has not been very engaged. In fact, the canonization of Firth and Harris during this time is paradoxical. On the one hand, it seems that they are invoked to lend theoretical authority to a field that struggles to lift its gaze from the latest state-of-the-art numbers \cite{manning_computational_2015, bender_climbing_2020}. Yet, the unspoken conclusion from the ascent of neural models and the language modeling revolution was that ``learning from data made linguistic theories irrelevant'' \cite[6295]{henderson_unstoppable_2020}. In other words, just as NLP seemed to lose interest with linguistic theory, it elevated two pioneering theoreticians to canonical status, but seemingly without engaging closely with their work. In fact, it often seems as if Firth and Harris are referenced in such a cavalier manner that it deflects attention from the field's general lack of engagement with linguistic theory. Meanwhile, Firth and Harris became figures who justify a relatively narrow conception of meaning, one that is predominantly intra-linguistic, without much to say about its usage in social life. 

This peculiar story has not been properly told. Though Léon discusses the contrast between Harris and Firth in the context of corpus linguistics (\citeyear{leon_empirical_2008}) and their influence on the history of computational linguistics (\citeyear{leon_automating_2021}), her work does not address the differences in their distributional theories and conceptions of ``context,'' nor the renewed and paradoxical significance of the two authors for language modeling. In our contribution, we emphasize the gap between the ideas of Firth and Harris as well as the insights a re-reading of their work offers for expanding the scope of computational semantics.

\section{Harris’s distributional structuralism}

Few linguists contributed more to linguistic theory than Zellig Harris (1909--1992), and not just by serving as Noam Chomsky's doctoral advisor. In fact, the two came to share little in common \cite{goldsmith_legacy_2005, nevin2010}. Whereas Chomsky’s generative grammar repositioned linguistics as a cognitive science seeking to understand, in so few words, the idealized mental representations and structures enabling language acquisition and production \cite[e.g.][]{chomsky_language_1972}, Harris’ radically distributional approach to language effectively elevated the natural language corpus as the sole starting point from which linguistic theory could arise (\citealp[1]{harris_methods_1951}; \citealp[2--3]{harris_form_1988}; \citealp[143--144]{nevin_computability_2002}).

This theory consisted of a linguistic structure segmentable into a finite set of formal objects characterized by constrained patterns of correspondence with one another \cite{harris_methods_1951, harris_distributional_1954, harris_theory_1991}. Such patterns of correspondence can be observed only in language-in-use, that is, in natural language corpora. In his foundational paper ``Distributional Structure,'' for example, Harris (\citeyear[156--157]{harris_distributional_1954}) provides a purely distributional account of how one might induce the semantic meanings of \textit{oculist}, \textit{eye-doctor}, and \textit{lawyer} from the partial (in the case of \textit{oculist} and \textit{lawyer}) or nearly complete (in the case of \textit{oculist} and \textit{eye-doctor}) overlap in their observed ``environments'' of use. This approach applies to other levels of linguistic analysis, such as morphophonemics \cite[e.g.,][155]{harris_distributional_1954}. Rather than producing a series of descriptive rules for the distribution of each phoneme, morpheme, or word, greater parsimony was sought by grouping these elements into structurally equivalent classes---categorized by their relationships as ``operators'' and ``arguments'' in Harris' later work (e.g. \citeyear{harris_mathematical_1968,harris_language_1988,harris_theory_1991})---sharing the same distributional rules, compounding elements in a hierarchical manner.

In this section we draw attention to three aspects of Harris's distributional linguistics: the relationship it posits between meaning and form; assumptions about heterogeneity among speakers of the same dialect; and the concept of sublanguages.

\subsection{Meaning and form}

A result of Harris' vision of a linguistics---concerned above all with the structural and probabilistic constraints governing the combination of formal elements---is that the discipline would be fully autonomous, not only from biology and psychology but even from semantics, phonetics, and logic, ``complete without intrusion of other features such as history or meaning'' (\citealt[146]{harris_distributional_1954}; \citealt[725--726]{goldsmith_legacy_2005}).\footnote{Indeed, as Jacqueline Léon notes (personal communication, April 18, 2022), even calling Harris' approach ``semantics'' is bit of an oxymoron. We elide a full discussion on the term since ``distributional semantics'' has become a commonplace phrase in NLP.} Harris's reasoning depended on the particular status he gave to linguistics among all the sciences. Taking language as its object of inquiry, linguistics lacks---unlike other sciences---a \textit{metalanguage} external to language, i.e., to its object of inquiry (\citealp[4--5]{harris_theory_1991}; \citealp[356]{nevin_minimalist_1993}). Even if some other symbolic system is used, ``those symbols will have to be defined ultimately in a natural language'' \cite[274]{harris_theory_1991}, as the surging demand for interpretability and explainability in NLP has made evident \cite[see e.g.][]{danilevsky_survey_2020}. Language is consequently not a ``code'' of ``forms'' that correlate with some meanings outside of it. It has no ``one-to-one'' conformity ``with some independently discoverable structure of meaning'' \cite[152]{harris_distributional_1954}. Instead, it is a system related to, but also independent from, thought (\citealp[383--384]{harris_theory_1991};  \citealp[361--363]{nevin_minimalist_1993}). While all human activity is meaningful, the particular meanings of language are \textit{constituted} by its form, not \textit{correlated} with it \cite[394]{nevin_minimalist_1993}. Meaning thus understood is about departures from equiprobability in the distribution of these constraints \cite[23]{harris_theory_1991}. These departures ``define a range of meaning for each morpheme, which includes its meaning in each occurrence''. Nonetheless, shared environments do not necessarily imply shared meanings: ``bumped into a pole can be said after a minor accident or after a chance meeting with an East European'' \citep[191]{harris_methods_1951}.

The ultimate goal of Harris's linguistic inquiry is to evaluate the efficiency of different grammars and their ability to model the statistical constraints imposed upon the distribution of different linguistic elements \cite[723--725]{goldsmith_legacy_2005}. Harris held that language was a ``detached pattern'' \cite[295]{harris_review_1941}---information that was public and socially transmissible and hence constitutive of new types of socially shared and conventional meaning (\citealp[342--345, 377--382]{harris_theory_1991}; \citealp[360, 365]{nevin_minimalist_1993})---and linguistics could at best discover different incomplete grammars \cite[31--36]{harris_theory_1991}. Though linguistics might provide insights about meaning and discourse, or about cultural practices, such findings would not bear directly on linguistics \textit{per se} \cite[725--726]{goldsmith_legacy_2005}. Indeed, while Harris acknowledged that our sense of word meaning is aided by ``extra-linguistic situational information,'' words ``beyond the immediate situation'' are ``on their own'' \cite[368]{harris_theory_1991}. However, Harris's method is not completely detached. Searching for a method to segment speech, he notes that the similarity of elements ``reduces ultimately to the similarity of sound segments under repetition,'' implemented through ``the pair test'' in which native speakers are asked to discriminate between sound segments \cite[158--159]{harris_distributional_1954}, producing an observational primitive that is ``more easily controlled than data on meaning'' \cite[e.g.][20]{harris_methods_1951}.

\subsection{Variation, or lack thereof}

Harris's view of language and linguistics, isolated from the vagaries of social interaction and variation, is obviously difficult to reconcile with a sociolinguistic perspective. As Harris writes in his \textit{Structural Linguistics} (\citeyear[9]{harris_methods_1951}), his approach is meant to describe a homogeneous dialect, which ``[i]n most cases...presents no problem, since the whole speech of the person or community shows dialectal consistency.'' Referencing this passage, sociolinguist and dialectology pioneer William Labov (\citeyear[5]{labov_social_2006}) argues that ``the inconsistency found in most New York City idiolects is so great that the first alternative of Harris is impossible, and the second implausible.''\footnote{New York City is not unique in this regard; it was merely the location of Labov's early pathbreaking work.} In other words, even at the level of the speaker, Harris’s idealized, unvarying idiolect does not hold up to empirical scrutiny. Rigorous consideration of factors that Harris would deem extra-linguistic (class, race, interactional roles, etc.) are indeed essential to produce a systematic description of linguistic structure \cite{labov_sociolinguistic_1972}. From the sociolinguistic perspective, Harris’s vision of a linguistic science fully isolated from the ``intrusion'' of non-verbal social life would never obtain the systemicity to which it aspired.

This sociolinguistic critique highlights fundamental limitations of Harris's perspective. Language is viewed primarily through the distributional restrictions imposed by convention, rather than by ``stylistic practice'' and the ways in which speakers ``make social-semiotic moves, reinterpreting variables and combining and recombining them in a continual process'' \cite[94]{eckert_three_2012}. We can study changes in discourse, as Harris himself did in an impressive volume on structures in immunological theory over time \cite{harris_form_1988}, but not how people make those changes, or indeed the way in which it is ``the variation itself that is systemic''  \cite[93]{deleuze_thousand_1987}. 

\subsection{Sublanguages}

As noted above, Harris' revival in the 1990s was driven by the new interest in ``corpus linguistics'' of large corpora, a research paradigm that was partly derived from Harris' notion of ``sublanguages'' \cite[]{leon_automating_2021}.
Harris introduced sublanguages in his book \textit{Mathematical Structures of Language}, defined as ``[c]ertain proper subsets of a language [which] may be closed under some or all of the operations defined in the language, and [which] thus constitute a sublanguage of it'' \citep[152]{harris_mathematical_1968}. A sublanguage is a set of sentences which are a subset of the sentences of the ``whole'' language. However, the grammatical constraints of the sublanguage are not necessarily those of the whole language; rather, their grammars intersect (\citealt[Ch 11]{harris_mathematical_1968}; \citealt[1]{kittredge_sublanguage_1982}). In application the term has come to refer primarily to the grammar and vocabulary unique to or characteristic of a particular professional or scientific field \cite[e.g.][]{harris_language_1988}, an influential concept for early information retrieval research \cite{sager_sublanguage_1975, sager_natural_1981}.\footnote{An important early figure in NLP, Sager received her PhD in Linguistics at the University of Pennsylvania and was directly influenced by Harris's work. See, e.g, Hirschman, Grishman, and Sager (\citeyear{hirschman1975grammatically}).} Harris believed that sublanguages could be neatly identified using the distributional methods of his general linguistic program.

\section{Firth's contextual semantics}

Something similar to the sociolinguistic critique of Harris could be articulated from a different perspective, namely, through the work of J.R. Firth. Firth (1890--1960)---professor of General Linguistics at the University College of London and the first holder of a chair in that subject in Britain---independently formulated a distributional theory of lexical semantics. However, unlike Harris, Firth refused to treat meaning separately from pragmatics, and words apart from their broader ``context of situation.'' \cite[205--208]{robins_contribution_1997}

Firth never published a fully articulated exposition of his general theory of language (\citealt[216]{robins_contribution_1997}; \citealt[180]{thomas_fifty_2011}) and today, all of his work is not only out of print but also mostly unavailable online. Not understood by ``the contemporary scientism'' of American descriptivist linguistics and its pioneers like Harris, Firth was mostly ignored on the other side of the Atlantic (\citealt[2]{firth_introduction_1968}; \citealt[280]{pandit_review_1970}). Unlike many of his American contemporaries, Firth did not draw mainly from cognitive psychology and logic---the latter of which Firth thought had ``taken the heart out of language'' \cite[186]{firth_introduction_1957}---but from the work of Polish anthropologist Bronisław Malinowski \cite[211]{robins_contribution_1997}. Here, we focus on the evolution of his thoughts on meaning and collocation, as well as his notions of \textit{context of situation} and \textit{restricted language}.

\subsection{Meaning by collocation}

To Firth, the purpose of linguistics is to ``study meaning in its own terms'' (\citealt[145]{firth_ethnographic_1968}; \citealt[289]{senis_contribution_2015}). The famous phrase about the company that words keep concerned a particular ``mode of meaning'': ``meaning by collocation'' \cite[194]{firth_modes_1957}. Anticipating vocabulary now ubiquitous in NLP, Firth thought that this level of meaning could be found by examining the ``habitual collocations'' of words and the ``word-material'' in which they are ``most characteristically embedded'' \cite[11--12]{firth_synopsis_1957}. Meaning by collocation was an abstraction of \textit{syntagmatic relations} (\citealt[444]{oyelaran_aspects_1967}) that went beyond ``mere juxtaposition,'' stating instead ``an order of \textit{mutual expectancy}'' and ``mutual prehension'' \cite[12]{firth_synopsis_1957}. While mutual expectancy could be understood similarly to Joos's (\citeyear{joos_description_1950}) conditional probabilities of occurrence or the concept of Pointwise Mutual Information \cite{fano_transmission_1961}, the notion of ``prehension'' originates in the work of philosopher and mathematician Alfred North Whitehead (\citeyear{whitehead_modes_1938}, \citeyear{whitehead_process_1957}; see also \citealt{butt_whiteheadian_2013}) and concerns the manner in which one entity grasps another and makes it part of its own experience (\citealt[12]{christian_interpretation_1959}; \citealt[136]{bryant_democracy_2011}). 

Drawing on Whitehead's ``modes of thought'' (\citeyear{whitehead_modes_1938}; see also \citealt[1812]{butt_firth_2001}, \citealt[28]{schonthal_firth_2019}), Firth advocated a type of ``polysystemic'' linguistic analysis that was interested in different, congruent modes of meaning, whether phonetic, phonological, syntactic, or semantic, but always situated in broader social context (\citealt[27, 30]{firth_synopsis_1957}
; \citealt[214]{robins_contribution_1997}). In stark contrast to Harris, Firth explicitly rejected any efforts to create ``unity in linguistics'' \cite[48]{firth_structural_1968} or one system of analysis. Citing the later Wittgenstein (\citeyear{wittgenstein_philosophical_1953}, \citealt[11]{firth_synopsis_1957}), Firth was mainly interested in the concrete use of language, reversing the schema of Ferdinand de Saussure (\citeyear{de_saussure_course_2011}) in which language  (\textit{la langue}) is a system ``external to and on a different plane from individual phenonema,'' including the concrete instances (\textit{la parole}) of language use (\citealt[400]{firth_semantics_1949}; \citealt[44--45]{firth_personality_1950}). While his final ideas matured considerably later, Firth initially articulated his ideas about semantics in two 1935 papers, one on semantics \cite{firth_technique_1935} and one on phonology \cite{firth_use_1935}, using the term ``contextual distribution'' in both. However, Firth (\citeyear[18]{firth_synopsis_1957}) ultimately disavowed this initial distributional theory---which was not too dissimilar from Harris's---as ``useful'' but inadequate to act as the ``main principle'' in a theory ``of structures involving the statement of the values of the elements of structure by reference to systems.''

By distinguishing between system (syntagm) and structure (paradigm), Firth wanted to highlight two operational principles necessary for meaning by collocation: 1) substitution within ``the same level of abstraction,'' and 2) commutation across different levels \cite[140]{robins_phonology_1953}. Only substitution that does not produce commutation in a sequence, indicates similarity of value or function (\citealt[5]{firth_synopsis_1957}; \citealt[23]{firth_linguistic_1968})\footnote{For further details, see examples provided by Bursill-Hall (\citeyear[132][]{bursill-hall_levels_1960}).}. Two words are only substitutable---and hence similar in function and meaning---if their values do not commute across a particular sentence. Substitutability, then, does not equal synonymy. Take, for example, the following two phrases containing a) prepositional and b) adverbial uses of the word ``by'':

\begin{itemize}[noitemsep]
\setlist{nolistsep}
\item[(a)] They go by night.
\item[(b)] They go by night after night.
\end{itemize}

Now, ``by'' could be replaced by the word ``past'' without commuting the meaning of the other words in (b). However, replacing ``by'' with ``past'' would commute with the rest of (a) in an impossible way \cite[23--24]{firth_linguistic_1968}. This demonstrates how substitution concerns the relationship between ``by'' and ``past'' as two elements at the same level of analysis---i.e., lexical units---but in order to account for commutation, we need to look beyond this level to other levels of abstraction.\footnote{A useful analogy might be the way in which BERT handles different aspects of language at different layers of the model \cite[e.g.]{tenney-EtAl:2019:ICLR}. However, no matter how large, a language model like BERT does not account for the context of situation.}

Firth's conception of collocation and his frequent nods to Whitehead were part of his ``monistic'' approach that rejected the division between mind and body (\citealt[2]{firth_synopsis_1957}; \citealt[5]{firth_introduction_1968}) and all the other dualities---language and thought, word and idea, signifier and signified, expression and content \cite[86]{firth_general_1951}---that characterized the structuralist linguistics of his time. He similarly rejected any notion of linguistics as ``a \textit{theory of universals} for \textit{general linguistic description}'' \cite[21]{firth_synopsis_1957}. Anticipating contemporary concerns about language diversity in NLP \cite[e.g.][]{bender_benderrule_2019}, Firth called for the Western scholar to ``de-Europeanize himself'' and the English scholar, due to the universal use of his language, to ``de-Anglicize himself'' (\citealt[96]{firth_descriptive_1968}; \citealt[274]{senis_contribution_2015}).

\subsection{Context and connection}

Diverging from structuralist linguistics, Firth suggested that a text should always be given a ``renewal of connection with experience'' \cite[29]{firth_synopsis_1957}. This notion of meaning was influenced by Malinowski, for whom Firth worked as an assistant early in his career \cite[346]{plug_j_2008} and from whom he borrowed the notion of ``context of situation''  (\citealt{firth_technique_1935}, \citealt[211]{robins_contribution_1997}). In Malinwoski's view, meaning was more than just a dyadic relationship between a word and its referent, ``a multidimensional and functional set of relations between the word in its sentence and the context of its occurrence'' \cite[35]{ardener_malinowski_1971}. However, while Malinowski's view on meaning was entirely functional and hyper-local, Firth employed the notion of ``context of situation'' as a necessary abstraction, not as a shorthand for things in themselves (\citealt[43]{firth_personality_1950}, \citealt[6]{firth_introduction_1968}). Context of situation is derived from an analytical choice, ``a set of categories in ordered relations abstracted from the life of man in the flux of events, from personality in society,'' \cite[30]{firth_synopsis_1957} prehending something of importance and bracketing the rest. It is, then, not necessarily about restricting the meaning of every utterance to a specific time and place, but about defining ``an abstract set of semantically relevant categories, abstracted from multitudes of actual situations, to which unique particulars could be referred.'' \cite[41--42]{ardener_malinowski_1971} Firth called for the linguist to focus on ``attested language text duly recorded'', accounting for a text's associated context of situation \textit{and} its interior relations. \cite[29--30]{firth_synopsis_1957}

Firth was famously opaque with the exact operationalization of his concepts, including context of situation, but he did provide a detailed list of the different contextual elements that a linguist should bring into relation during analysis \cite[43]{firth_personality_1950}. These include the relevant features of participants (persons, personalities); their verbal and non-verbal actions; the relevant objects; and the effects of verbal action.

During Firth's lifetime, the most thorough work that put his notion of context of situation to work was an ethnographic study by his student T.F. Mitchell in former Cyreneica (today Libya) on the language of buying and selling at the local markets of different cities and villages in the region. For Mitchell (\citeyear[32--33]{mitchell_language_1957}), contexts that might ``correlate'' with particular types of text included: the spatio-temporal situation of persons in the context; the activities of participants; the attitudes of the participants; their ``personalities'' such as specific trade of profession, geographical and class origins, educational standard, inter-relationship, and so on.

It is worth noting that both Mitchell and M.A.K. Halliday---Firth's student who synthesized much of his theory---used words such as ``correlation,'' ``inference,'' and ``prediction'' to describe the relationship between a text and its situational context, implying that a statistical extension of their approaches would not be completely unreasonable. In fact, Halliday himself suggested as much, when he in the early 1990s made efforts to bridge his branch of linguistics with the nascent field of corpus linguistics \cite{halliday_corpus_1991}.

In conclusion: Firth's famous quote itself refers to collocation, while his notion of ``context'' implies something much broader, ``the whole conceptual meaning'' \cite[11]{firth_synopsis_1957}. Context is the ground against which the figure of the text must be understood, no matter (per Harris) how ``detached'' its pattern might be \cite[e.g.][]{auer_context_1996}. Without context, collocation captures only one narrow ``mode of meaning.''

\subsection{Restricted languages}

Like Harris, Firth's revival in connection with 1990s corpus linguistics was related to his attempts to respond to practical needs of empirical research.
Expanding upon his functional understanding of language, Firth developed his notion of ``restricted languages'' in the 1950s (\citealt[7]{leon_linguistic_2007}). In a posthumously published essay, he describes social actors as ``collect[ing] a varied repertory of interlocking roles'' corresponding to a ``constellation of restricted languages''  \cite[207]{firth_treatment_1968}. As people shift between locally contextualized roles, they draw upon their ``repertory'' of restricted languages with specialized vocabulary and discursive styles that both reflect and constitute these contexts. Thus one might speak of a ``restricted language of science, sport, defense, industry, aviation, military services, commerce, law and civil administration, politics, literature, etc.'' \cite[261]{leon_empirical_2008}. As such, the concept of restricted language is now generally seen as a precursor to the concept of ``register,'' which was taken up by subsequent sociolinguists and linguistic anthropologists  \cite[e.g.][]{halliday_users_1968, gumperz_directions_1972, agha_voice_2005}.

In proposing restricted languages as the proper object of descriptive linguistic analysis, Firth was making a broader theoretical point against, on the one hand, ``the monosystemic view of language'' of neo-Bloomfieldians like Harris and ``pointless discussions on metalanguage'' on the other, for metalanguage could be reanalyzed as a ``restricted language of linguistics'' itself \cite[9]{leon_linguistic_2007}. Simply put, a descriptive linguistics which privileges restricted languages also necessarily privileges contexts of situation as an essential dimension of variation that allows social meaning to inhere in language.

\section{Discussion: Words in mixed company}

Often cited, together or separately, to justify a distributional appraoch to semantics, Firth and Harris nonetheless offer differing views on language and meaning. Harris offers us a rigorous formalism that treats language as a ``detached pattern''---not a ``code'', but a particular system of meaning. Firth, by comparison, left a much more scattered legacy that was only systematized by his students. Firth and Harris shared a concern about the lack of an external metalanguage of linguistics, but drew different conclusions from it. If Harris responded to this conundrum by creating one hierarchically organized system without intrusion from extra-linguistic factors, Firth called for an investigation of language as a ``spectrum'' \citep[76]{firth_general_1951} with different modes of meaning that had to be addressed through multiple levels of analysis---starting with the context of situation and proceeding from there to decide which other levels are relevant \citep[44]{firth_personality_1950}.  
Firth's distributional theory has been described unfavourably as based on frequent co-occurrence, in contrast to the recursive dependencies developed by Harris \cite[205]{nevin_contextual_2002}. For Harris, the meaning of a word depends on its set \cite[e.g.][17]{harris_theory_1991} such that, for example, the words ``divide'' and ``multiply'' operate on the word ``cell'' (and vice versa) in the same way, producing essentially the same meaning (\citealt[62]{harris_language_1988}). However, Firth's final method of substitution and commutation also establishes complex, multidimensional criteria for distributional contrast as well as a framework for understanding polysemy. Though less formalized and less obviously recursive than Harris', Firth's approach can, arguably, also be read as treating linguistic elements as operators and arguments defined by their sets in a complex hierarchy (\citealp[44]{firth_personality_1950}; \citealp[76]{firth_general_1951} ``at a series of congruent levels'' \citep[29--30]{firth_synopsis_1957} with different ``bands of abstraction'' \citep[49]{firth_structural_1968}, \textit{including} the extra-linguistic context of situation. Harris and Firth both understand any linguistic analysis as incomplete, for Harris always a pursuit of the least description'' \citep[3]{harris_language_1988}---i.e., best ``grammar'' or model---for Firth always grounded in the social construction of facts, without any possibility of ``complete axiomatization'' \citep[44--45]{firth_structural_1968}.

From the perspective of empirical work, especially decades after their time, Firth and Harris also share similarities. Both rejected the mentalism that was so prevalent during their time. They were both revived as empiricist originators during the rise of statistical learning in the 1990s, and their respective work on restricted languages and sublanguages largely conflated in service of the practical concerns of corpus linguistics \cite{leon_empirical_2008}. Their theories both included in what we might call a ``reality principle,'' a final arbiter of meaning outside of form: the pair test for Harris, the context of situation for Firth. The former grounds linguistics in the smallest possible unit of analysis as understood by the native speaker, the latter in social actions and objects.

In light of recent calls to extend the ``world scope'' \cite{bisk_experience_2020} of NLP and to move towards pragmatic notions of meaning, it might make sense to balance Harris's formalism and Firth's pluralism. Though Firth warned us against overextending linguistics, he was generous with the company that words could keep. They mingled with each other, but also with events, objects, people, and indexical features such as time and space. And if NLP is ready to move beyond the corpus, then even Harris might acknowledge that when modeling language in ``the immediate situation''---whether in online interactions or face-to-face communication---words are \textit{not} on their own, that to judge the meaning of a combination of words, we can summon ``the aid of some of the extra-linguistic situational information'' \cite[368]{harris_theory_1991}. In the following subsections, we consider two ways in which NLP is already doing this, in order to highlight some already existing strategies for broader contextualization. We call these strategies ``comparative stratification'' and ``syntagmatic extension.''

\subsection{Comparative stratification}

Corpus linguistics emerges from the question of what kind of company words keep, depending on their context. The issue of context was motivated by the introduction of corpus linguistics both for students of Firth---who considered restricted languages as a way of handling context---and for Harris's sublanguages, which were ``contextually situated and suitable for being processed automatically'' \cite[149--150]{leon_automating_2021}. However, beyond just studying the restricted corpus, we might also consider the ways in which large datasets can be ``stratified,'' systematically dividing them into sub-corpora that are studied in relation to each other. Here, the company that words keep among each other is \textit{limited} for analytical purposes, but in a manner that implies a relationship between that ``company'' and situational context.

Diachronic embeddings are especially representative of this approach. By stratifying timestamped data into a number of intervals, training separate models for each and then aligning the embeddings using either ``second-order embeddings'' or methods such as linear transformations \cite{kutuzov_diachronic_2018}, the analyst can effectively represent a temporal ``context of situation.'' This approach works with both static \cite{hamilton_diachronic_2016} and contextual embeddings \cite{martinc_leveraging_2020}. While variants of this approach are most commonly used to study semantic shifts \cite[e.g.,][]{garg_word_2018, kozlowski_geometry_2019, mendelsohn2020framework}, it could plausibly be used to stratify a dataset according to other variables such as space \cite[e.g.][]{bamman_distributed_2014,gong_enriching_2020}, online communities \cite{lucy2021characterizing}, persons \cite{yao2020employing}, or domains \cite{spinde2021identification}. Words still only keep the company of one another, but by limiting their company we implicitly introduce other participants in the analysis.\footnote{In a very broad sense, the trend of pretraining large language models and then finetuning them on specific datasets is of course also an admission of the importance of ``context of situation.''}

\subsection{Syntagmatic extension}

Recall that for Firth, meaning by collocation and considerations of a ``typical context of situation'' \cite[44]{firth_personality_1950} were exercises in abstraction, with collocation being an abstraction at the syntagmatic level. Instead of restricting the company words keep, we might follow Firth's recommendation to consider them in wider company ``of the same abstract nature'' \cite[7]{firth_personality_1950}. In vector semantics, this would imply that we explicitly introduce different contextual factors in the \textit{same} vector space with our words, endowing them all with ontological equality.

The paragraph vectors introduced by Le and Mikolov (\citeyear{le_distributed_2014}) as an extension of the earlier Skip-gram algorithm \cite{mikolov_distributed_2013} are representative for this approach. In practice, this method extends the syntagmatic chain of words by introducing a vector for the document as a new paradigmatic element. In principle, this type of ``global context'' \cite{grbovic_real-time_2018} could be anything and include several paradigmatic elements, as we can see in the research on multi-modal embeddings \cite[e.g.][]{baroni_grounding_2016} and generative modeling \cite[e.g.][]{ramesh_hierarchical_2022}.\footnote{Firth himself (\citeyear[26]{firth_synopsis_1957}) recommended accompanying word definitions and collocational information with pictures.} Models have been developed that include demographic \cite{garimella2017demographic} or persona \cite{li2016persona} vectors in the embedding space, such that intra-textual relations are accompanied by information about speakers' social categories. However, implemented with static embeddings and without some additional grammar restrictions, these context vectors essentially add only a ``bag of contexts.'' For static embeddings, additional grammar constraints could be introduced, as was done in research on Point-of-Interest (POI) data in the the domain of geosemantics, where researchers constrained contextual vectors using spatial variograms \cite{yan_itdl_2017}. Beyond static embeddings, large-language models and their dynamic embeddings could either be pretrained (with the appropriate dataset) or finetuned on data with text associated with different contextual variables. This would realize the proposal that Halliday made in the early 1990s when he suggested an extension of the language modeling schema from the early work of Shannon, to a model with ``global probabilities, those of the grammar of English, and the locally conditioned probabilities, those of this or that particular register'' \cite[37]{halliday_corpus_1991}.

\section{Conclusions}

This paper revisited the theories of the two most well-known progenitors of the distributional approach to meaning in NLP. Recognizing the open question of how to bring NLP beyond the corpus, we offer a thorough account of the two distributional theories that are most often invoked to justify the modeling of meaning through departures from randomness in the company that words keep. Comparing the work of Harris and Firth---who both published their major work before the rise of the internet and its corpora---we find two distinct theories of distribution: one formal and mathematical, treating language as a particular type of detached information, another more schematic and anthropological, treating language as a functional spectrum which always emanates from a particular context of situation. The legacies of both Firth and Harris can be seen in the current paradigm of corpus linguistics, but in the domain of distributional semantics, it is Harris's ethos that dominates, despite Firth providing its most famous tagline. 

Moving forward, we suggest that semantic modeling take more inspiration from Firth, and consider the context of situation and the wide variety of company that words can keep as crucial sites of innovation for the field. Doing so may not involve following a finite set of steps or flowchart. Rather, we humbly suggest that the field may be enriched by thoughtful and creative re-engagements with the intellectual traditions from which it has historically drawn. This does not imply abandoning the rigor provided by Harris. On the contrary, we find that Firth and Harris would probably have agreed that any model or ``grammar'' is inevitably incomplete and partial. No universal model is possible, despite the large-language modeling fervor, nor will there be one theory of language to guide us. There are only the partial perspectives and the inevitable choice of adapting one.

\section*{Acknowledgements}

We would like to thank Dan Jurafsky and Jacqueline Léon for their gracious and helpful comments which greatly improved the paper. Mikael Brunila is funded by the Kone Foundation, and Jack LaViolette is funded by the Mellon Foundation. We thank them for their generous support.

\bibliography{anthology,custom}

\begin{thebibliography}{143}
\expandafter\ifx\csname natexlab\endcsname\relax\def\natexlab#1{#1}\fi

\bibitem[{Agha(2005)}]{agha_voice_2005}
Asif Agha. 2005.
\newblock \href
  {https://anthrosource.onlinelibrary.wiley.com/doi/abs/10.1525/jlin.2005.15.1.38?casa_token=YOoxKKR7e3EAAAAA:XhD2GjhG5fWuisQ4twVW3Anlgvx4rIO48pPaYj10Ntu1Wch-GgYwmURZVrVFWA9zaJMpDsRQbjwtTWmt}
  {Voice, footing, enregisterment}.
\newblock \emph{Journal of Linguistic Anthropology}, 15(1):38--59.

\bibitem[{Auer(1996)}]{auer_context_1996}
Peter Auer. 1996.
\newblock \href {https://raco.cat/index.php/LinksLetters/article/view/49842}
  {From context to contextualization}.
\newblock \emph{Links \& Letters}, 3(1):11--28.

\bibitem[{Bahdanau et~al.(2015)Bahdanau, Cho, and Bengio}]{bahdanau2015neural}
Dzmitry Bahdanau, Kyunghyun Cho, and Yoshua Bengio. 2015.
\newblock \href {http://arxiv.org/abs/1409.0473} {Neural machine translation by
  jointly learning to align and translate}.
\newblock In \emph{3rd International Conference on Learning Representations,
  {ICLR} 2015, May 7-9, 2015, Conference Track Proceedings}, pages 1--15, San
  Diego, CA, USA.

\bibitem[{Bamman et~al.(2014)Bamman, Dyer, and Smith}]{bamman_distributed_2014}
David Bamman, Chris Dyer, and Noah~A. Smith. 2014.
\newblock \href {https://doi.org/10.3115/v1/P14-2134} {Distributed
  representations of geographically situated language}.
\newblock In \emph{Proceedings of the 52nd {Annual} {Meeting} of the
  {Association} for {Computational} {Linguistics} ({Volume} 2: {Short}
  {Papers})}, pages 828--834, Baltimore, MD, USA. Association for Computational
  Linguistics.

\bibitem[{Baroni(2016)}]{baroni_grounding_2016}
Marco Baroni. 2016.
\newblock \href {https://doi.org/10.1111/lnc3.12170} {Grounding distributional
  semantics in the visual world}.
\newblock \emph{Language and Linguistics Compass}, 10(1):3--13.

\bibitem[{Bender(2019)}]{bender_benderrule_2019}
Emily~M. Bender. 2019.
\newblock \href
  {https://thegradient.pub/the-benderrule-on-naming-the-languages-we-study-and-why-it-matters/}
  {The \#{BenderRule}: {On} naming the languages we study and why it matters}.
\newblock \emph{The Gradient}.

\bibitem[{Bender et~al.(2021)Bender, Gebru, McMillan-Major, and
  Shmitchell}]{bender_dangers_2021}
Emily~M. Bender, Timnit Gebru, Angelina McMillan-Major, and Shmargaret
  Shmitchell. 2021.
\newblock \href {https://doi.org/10.1145/3442188.3445922} {On the dangers of
  stochastic parrots: Can language models be too big?}
\newblock In \emph{Proceedings of the 2021 {ACM} {Conference} on {Fairness},
  {Accountability}, and {Transparency}}, {FAccT} '21, pages 610--623, New York,
  NY, USA. Association for Computing Machinery.

\bibitem[{Bender and Koller(2020)}]{bender_climbing_2020}
Emily~M. Bender and Alexander Koller. 2020.
\newblock \href {https://doi.org/10.18653/v1/2020.acl-main.463} {Climbing
  towards {NLU}: On meaning, form, and understanding in the {Age} of {Data}}.
\newblock In \emph{Proceedings of the 58th {Annual} {Meeting} of the
  {Association} for {Computational} {Linguistics}}, pages 5185--5198, Online.
  Association for Computational Linguistics.

\bibitem[{Bengio et~al.(2003)Bengio, Ducharme, Vincent, and
  Janvin}]{bengio_neural_2003}
Yoshua Bengio, Réjean Ducharme, Pascal Vincent, and Christian Janvin. 2003.
\newblock \href {https://dl.acm.org/doi/10.5555/944919.944966} {A neural
  probabilistic language model}.
\newblock \emph{The Journal of Machine Learning Research}, 3:1137--1155.

\bibitem[{Bisk et~al.(2020)Bisk, Holtzman, Thomason, Andreas, Bengio, Chai,
  Lapata, Lazaridou, May, Nisnevich, Pinto, and Turian}]{bisk_experience_2020}
Yonatan Bisk, Ari Holtzman, Jesse Thomason, Jacob Andreas, Yoshua Bengio, Joyce
  Chai, Mirella Lapata, Angeliki Lazaridou, Jonathan May, Aleksandr Nisnevich,
  Nicolas Pinto, and Joseph Turian. 2020.
\newblock \href {https://doi.org/10.18653/v1/2020.emnlp-main.703} {Experience
  grounds language}.
\newblock In \emph{Proceedings of the 2020 {Conference} on {Empirical}
  {Methods} in {Natural} {Language} {Processing} ({EMNLP})}, pages 8718--8735,
  Online. Association for Computational Linguistics.

\bibitem[{Blei et~al.(2002)Blei, Ng, and Jordan}]{blei_latent_2002}
David~M. Blei, Andrew~Y. Ng, and Michael~I. Jordan. 2002.
\newblock \href
  {http://papers.nips.cc/paper/2070-latent-dirichlet-allocation.pdf} {Latent
  {Dirichlet} {Allocation}}.
\newblock In T.~G. Dietterich, S.~Becker, and Z.~Ghahramani, editors,
  \emph{Advances in {Neural} {Information} {Processing} {Systems} 14}, pages
  601--608. MIT Press.

\bibitem[{Blodgett et~al.(2020)Blodgett, Barocas, Daumé~III, and
  Wallach}]{blodgett_language_2020}
Su~Lin Blodgett, Solon Barocas, Hal Daumé~III, and Hanna Wallach. 2020.
\newblock \href {http://arxiv.org/abs/2005.14050} {Language (technology) is
  power: {A} critical survey of "bias" in {NLP}}.
\newblock In \emph{Proceedings of the 58th Annual Meeting of the Association
  for Computational Linguistics}, pages 5454--5476, Online. Association for
  Computational Linguistics.

\bibitem[{Bojanowski et~al.(2017)Bojanowski, Grave, Joulin, and
  Mikolov}]{bojanowski_enriching_2017}
Piotr Bojanowski, Edouard Grave, Armand Joulin, and Tomas Mikolov. 2017.
\newblock \href {https://doi.org/10.1162/tacl_a_00051} {Enriching word vectors
  with subword information}.
\newblock \emph{Transactions of the Association for Computational Linguistics},
  5:135--146.

\bibitem[{Bourdieu(1984)}]{bourdieu_distinction_1984}
Pierre Bourdieu. 1984.
\newblock \href {https://www.hup.harvard.edu/catalog.php?isbn=9780674212770}
  {\emph{Distinction: {A} social critique of the judgement of taste}}.
\newblock Harvard University Press, Cambridge, MA.

\bibitem[{Bruni et~al.(2014)Bruni, Tran, and Baroni}]{bruni_multimodal_2014}
E.~Bruni, N.~K. Tran, and M.~Baroni. 2014.
\newblock \href {https://doi.org/10.1613/jair.4135} {Multimodal distributional
  semantics}.
\newblock \emph{Journal of Artificial Intelligence Research}, 49:1--47.

\bibitem[{Bryant(2011)}]{bryant_democracy_2011}
Levi~R. Bryant. 2011.
\newblock \href {https://library.oapen.org/handle/20.500.12657/33908}
  {\emph{The democracy of objects}}.
\newblock Open Humanities Press, Ann Arbor, MI, USA.

\bibitem[{Burgess(1998)}]{burgess1998simple}
Curt Burgess. 1998.
\newblock \href {https://link.springer.com/article/10.3758/BF03200643} {From
  simple associations to the building blocks of language: Modeling meaning in
  memory with the {HAL} model}.
\newblock \emph{Behavior Research Methods, Instruments, \& Computers},
  30(2):188--198.

\bibitem[{Bursill-Hall(1960)}]{bursill-hall_levels_1960}
G.~L. Bursill-Hall. 1960.
\newblock \href {https://doi.org/10.1017/S0008413100019125} {Levels analysis:
  {J}. {R}. {Firth}’s theories of linguistic analysis}.
\newblock \emph{Canadian Journal of Linguistics/Revue canadienne de
  linguistique}, 6(2):124--135.

\bibitem[{Butt(2001)}]{butt_firth_2001}
David~G. Butt. 2001.
\newblock \href {https://doi.org/10.1515/9783110167351.2.31.1806} {Firth,
  {Halliday} and the development of systemic functional theory}.
\newblock In Sylvain Auroux, E.~F.~K. Koerner, Hans-Josef Niederehe, and Kees
  Versteegh, editors, \emph{History of the language sciences}, volume~2 of
  \emph{Handbücher zur {Sprach}- und {Kommunikationswissenschaft} /
  {Handbooks} of {Linguistics} and {Communication} {Science} ({HSK})}, pages
  1806--1838. De Gruyter, Berlin.

\bibitem[{Butt(2013)}]{butt_whiteheadian_2013}
David~G. Butt. 2013.
\newblock \href {https://doi.org/10.1515/9783110333299.2.21} {Whiteheadian and
  functional linguistics}.
\newblock In Michael Weber, editor, \emph{Handbook of Whiteheadian Process
  Thought}, pages 21--32. De Gruyter, Berlin.

\bibitem[{Butt(2019)}]{schonthal_firth_2019}
David~G. Butt. 2019.
\newblock \href {https://doi.org/10.1017/9781316337936.003} {Firth and the
  origins of systemic functional linguistics: {Process}, pragma, and
  polysystem}.
\newblock In David Schönthal, Geoff Thompson, Lise Fontaine, and Wendy~L.
  Bowcher, editors, \emph{The {Cambridge} {Handbook} of {Systemic} {Functional}
  {Linguistics}}, Cambridge {Handbooks} in {Language} and {Linguistics}, pages
  11--34. Cambridge University Press, Cambridge, UK.

\bibitem[{Chomsky(1972)}]{chomsky_language_1972}
Noam Chomsky. 1972.
\newblock \href {https://www.worldcat.org/title/language-and-mind/oclc/265760}
  {\emph{Language and mind}}.
\newblock Harcourt Brace Jovanovich, New York, NY, USA.

\bibitem[{Chowdhery et~al.(2022)Chowdhery, Narang, Devlin, Bosma, Mishra,
  Roberts, Barham, Chung, Sutton, Gehrmann, Schuh, Shi, Tsvyashchenko, Maynez,
  Rao, Barnes, Tay, Shazeer, Prabhakaran, Reif, Du, Hutchinson, Pope, Bradbury,
  Austin, Isard, Gur-Ari, Yin, Duke, Levskaya, Ghemawat, Dev, Michalewski,
  Garcia, Misra, Robinson, Fedus, Zhou, Ippolito, Luan, Lim, Zoph, Spiridonov,
  Sepassi, Dohan, Agrawal, Omernick, Dai, Pillai, Pellat, Lewkowycz, Moreira,
  Child, Polozov, Lee, Zhou, Wang, Saeta, Diaz, Firat, Catasta, Wei,
  Meier-Hellstern, Eck, Dean, Petrov, and Fiedel}]{chowdhery_palm_2022}
Aakanksha Chowdhery, Sharan Narang, Jacob Devlin, Maarten Bosma, Gaurav Mishra,
  Adam Roberts, Paul Barham, Hyung~Won Chung, Charles Sutton, Sebastian
  Gehrmann, Parker Schuh, Kensen Shi, Sasha Tsvyashchenko, Joshua Maynez,
  Abhishek Rao, Parker Barnes, Yi~Tay, Noam Shazeer, Vinodkumar Prabhakaran,
  Emily Reif, Nan Du, Ben Hutchinson, Reiner Pope, James Bradbury, Jacob
  Austin, Michael Isard, Guy Gur-Ari, Pengcheng Yin, Toju Duke, Anselm
  Levskaya, Sanjay Ghemawat, Sunipa Dev, Henryk Michalewski, Xavier Garcia,
  Vedant Misra, Kevin Robinson, Liam Fedus, Denny Zhou, Daphne Ippolito, David
  Luan, Hyeontaek Lim, Barret Zoph, Alexander Spiridonov, Ryan Sepassi, David
  Dohan, Shivani Agrawal, Mark Omernick, Andrew~M. Dai,
  Thanumalayan~Sankaranarayana Pillai, Marie Pellat, Aitor Lewkowycz, Erica
  Moreira, Rewon Child, Oleksandr Polozov, Katherine Lee, Zongwei Zhou, Xuezhi
  Wang, Brennan Saeta, Mark Diaz, Orhan Firat, Michele Catasta, Jason Wei,
  Kathy Meier-Hellstern, Douglas Eck, Jeff Dean, Slav Petrov, and Noah Fiedel.
  2022.
\newblock \href {http://arxiv.org/abs/2204.02311} {{PaLM}: {Scaling} language
  modeling with pathways}.
\newblock \emph{arXiv:2204.02311 [cs]}.
\newblock ArXiv: 2204.02311.

\bibitem[{Christian(1959)}]{christian_interpretation_1959}
William~A. Christian. 1959.
\newblock \href
  {https://www.cambridge.org/core/journals/philosophy/article/abs/an-interpretation-of-whiteheads-metaphysics-by-christian-william-a-yale-university-press-london-oxford-university-press-1959-pp-419-price-48s/D1437FFE5E1952DB75912024ACD71887}
  {\emph{An interpretation of Whitehead's metaphysics}}.
\newblock Yale University Press, New Haven, CT, USA.

\bibitem[{Church and Liberman(2021)}]{church2021future}
Kenneth Church and Mark Liberman. 2021.
\newblock \href
  {https://www.frontiersin.org/articles/10.3389/frai.2021.625341/full} {The
  future of computational linguistics: On beyond alchemy}.
\newblock \emph{Frontiers in Artificial Intelligence}, 4:1--18.

\bibitem[{Church and Hanks(1989)}]{church_word_1989}
Kenneth~W. Church and Patrick Hanks. 1989.
\newblock \href {https://doi.org/10.3115/981623.981633} {Word association
  norms, mutual information, and lexicography}.
\newblock In \emph{27th {Annual} {Meeting} of the {Association} for
  {Computational} {Linguistics}}, pages 76--83, Vancouver, British Columbia,
  Canada. Association for Computational Linguistics.

\bibitem[{Church and Mercer(1993)}]{church_introduction_1993}
Kenneth~W. Church and Robert~L. Mercer. 1993.
\newblock \href {https://aclanthology.org/J93-1001} {Introduction to the
  {Special} {Issue} on {Computational} {Linguistics} {Using} {Large}
  {Corpora}}.
\newblock \emph{Computational Linguistics}, 19(1):1--24.

\bibitem[{Danilevsky et~al.(2020)Danilevsky, Qian, Aharonov, Katsis, Kawas, and
  Sen}]{danilevsky_survey_2020}
Marina Danilevsky, Kun Qian, Ranit Aharonov, Yannis Katsis, Ban Kawas, and
  Prithviraj Sen. 2020.
\newblock \href {https://www.aclweb.org/anthology/2020.aacl-main.46} {A
  {Survey} of the {State} of {Explainable} {AI} for {Natural} {Language}
  {Processing}}.
\newblock In \emph{Proceedings of the 1st {Conference} of the {Asia}-{Pacific}
  {Chapter} of the {Association} for {Computational} {Linguistics} and the 10th
  {International} {Joint} {Conference} on {Natural} {Language} {Processing}},
  pages 447--459, Suzhou, China. Association for Computational Linguistics.

\bibitem[{de~Saussure(1916/2011)}]{de_saussure_course_2011}
Ferdinand de~Saussure. 1916/2011.
\newblock \href
  {http://cup.columbia.edu/book/course-in-general-linguistics/9780231157261}
  {\emph{Course in general linguistics}}.
\newblock Columbia University Press, New York, NY, USA.

\bibitem[{Deerwester et~al.(1989)Deerwester, Dumais, Furnas, Harshman,
  Landauer, Lochbaum, and Streeter}]{deerwester_computer_1989}
Scott~C. Deerwester, Susan~T. Dumais, George~W. Furnas, Richard~A. Harshman,
  Thomas~K. Landauer, Karen~E. Lochbaum, and Lynn~A. Streeter. 1989.
\newblock \href {https://patents.google.com/patent/US4839853A/en} {Computer
  information retrieval using latent semantic structure. {US patent
  US4839853A}}.

\bibitem[{Deerwester et~al.(1990)Deerwester, Dumais, Furnas, Landauer, and
  Harshman}]{deerwester_indexing_1990}
Scott~C. Deerwester, Susan~T. Dumais, George~W. Furnas, Thomas~K. Landauer, and
  Richard~A. Harshman. 1990.
\newblock \href
  {https://doi.org/10.1002/(SICI)1097-4571(199009)41:6<391::AID-ASI1>3.0.CO;2-9}
  {Indexing by latent semantic analysis}.
\newblock \emph{Journal of the American Society for Information Science},
  41(6):391--407.

\bibitem[{Deleuze and Guattari(1987)}]{deleuze_thousand_1987}
Gilles Deleuze and Félix Guattari. 1987.
\newblock \href
  {https://www.upress.umn.edu/book-division/books/a-thousand-plateaus} {\emph{A
  thousand plateaus: {Capitalism} and Schizophrenia}}.
\newblock University of Minnesota Press, Minneapolis, MN, USA.

\bibitem[{Devlin et~al.(2019)Devlin, Chang, Lee, and
  Toutanova}]{devlin_bert_2019}
Jacob Devlin, Ming-Wei Chang, Kenton Lee, and Kristina Toutanova. 2019.
\newblock \href {https://doi.org/10.18653/v1/N19-1423} {{BERT}: {Pre}-training
  of {Deep} {Bidirectional} {Transformers} for language understanding}.
\newblock In \emph{Proceedings of the 2019 {Conference} of the {North}
  {American} {Chapter} of the {Association} for {Computational} {Linguistics}:
  {Human} {Language} {Technologies}, {Volume} 1 ({Long} and {Short} {Papers})},
  pages 4171--4186, Minneapolis, MN, USA. Association for Computational
  Linguistics.

\bibitem[{Eckert(2008)}]{eckert_variation_2008}
Penelope Eckert. 2008.
\newblock \href {https://doi.org/10.1111/j.1467-9841.2008.00374.x} {Variation
  and the indexical field}.
\newblock \emph{Journal of Sociolinguistics}, 12(4):453--476.

\bibitem[{Eckert(2012)}]{eckert_three_2012}
Penelope Eckert. 2012.
\newblock \href {https://doi.org/10.1146/annurev-anthro-092611-145828} {Three
  waves of variation study: The emergence of meaning in the study of
  sociolinguistic variation}.
\newblock \emph{Annual Review of Anthropology}, 41(1):87--100.

\bibitem[{Eisenstein(2019)}]{eisenstein_introduction_2019}
Jacob Eisenstein. 2019.
\newblock \href
  {https://mitpress.mit.edu/books/introduction-natural-language-processing}
  {\emph{Introduction to natural language processing}}.
\newblock MIT Press, Cambridge, MA, USA.

\bibitem[{Fano(1961)}]{fano_transmission_1961}
Robert~M. Fano. 1961.
\newblock \href {https://mitpress.mit.edu/books/transmission-information}
  {\emph{Transmission of information: {A} statistical theory of
  communication}}.
\newblock MIT Press, Cambridge, MA, USA.

\bibitem[{Firth(1935{\natexlab{a}})}]{firth_technique_1935}
J.R. Firth. 1935{\natexlab{a}}.
\newblock \href
  {https://doi.org/https://doi.org/10.1111/j.1467-968X.1935.tb01254.x} {The
  technique of semantics}.
\newblock \emph{Transactions of the Philological Society}, 34(1):36--73.

\bibitem[{Firth(1935{\natexlab{b}})}]{firth_use_1935}
J.R. Firth. 1935{\natexlab{b}}.
\newblock \href {https://doi.org/10.1080/00138383508596629} {The use and
  distribution of certain {English} sounds}.
\newblock \emph{English Studies}, 17(1-6):8--18.

\bibitem[{Firth(1949)}]{firth_semantics_1949}
J.R. Firth. 1949.
\newblock \href {https://doi.org/10.1016/0024-3841(49)90085-6} {The semantics
  of linguistic science}.
\newblock \emph{Lingua}, 1:393--404.

\bibitem[{Firth(1950)}]{firth_personality_1950}
J.R. Firth. 1950.
\newblock \href {https://doi.org/10.1111/j.1467-954X.1950.tb02460.x}
  {Personality and language in society}.
\newblock \emph{The Sociological Review}, a42(1):37--52.

\bibitem[{Firth(1951)}]{firth_general_1951}
J.R. Firth. 1951.
\newblock \href {https://doi.org/10.1111/j.1467-968X.1951.tb00249.x} {General
  linguistics and descriptive grammar}.
\newblock \emph{Transactions of the Philological Society}, 50(1):69--87.

\bibitem[{Firth(1957{\natexlab{a}})}]{firth_introduction_1957}
J.R. Firth. 1957{\natexlab{a}}.
\newblock \href {https://ci.nii.ac.jp/naid/10020680394/} {Introduction}.
\newblock In J.R. Firth, editor, \emph{Studies in {Linguistic} {Analysis}}.
  Basil Blackwell, Oxford, UK.

\bibitem[{Firth(1957{\natexlab{b}})}]{firth_modes_1957}
J.R. Firth. 1957{\natexlab{b}}.
\newblock \href
  {https://www.worldcat.org/title/papers-in-linguistics-1934-1951/oclc/32594232}
  {Modes of meaning}.
\newblock In \emph{Papers in {Linguistics}, 1934-1951}, pages 190--215. Oxford
  University Press, London, UK.

\bibitem[{Firth(1957{\natexlab{c}})}]{firth_synopsis_1957}
J.R. Firth. 1957{\natexlab{c}}.
\newblock \href {https://ci.nii.ac.jp/naid/10020680394/} {A synopsis of
  linguistic theory, 1930-1955}.
\newblock In J.R. Firth, editor, \emph{Studies in {Linguistic} {Analysis}}.
  Basil Blackwell, Oxford, UK.

\bibitem[{Firth(1968{\natexlab{a}})}]{firth_descriptive_1968}
J.R. Firth. 1968{\natexlab{a}}.
\newblock \href {http://125.22.75.155:8080/handle/123456789/12198} {Descriptive
  linguistics and the study of {English}}.
\newblock In Frank~R. Palmer, editor, \emph{Selected papers of {J}.{R}. {Firth}
  (1952-59)}, pages 96--113. Longman and Indiana University Press, London, UK,
  and Bloomington, IN, USA.

\bibitem[{Firth(1968{\natexlab{b}})}]{firth_ethnographic_1968}
J.R. Firth. 1968{\natexlab{b}}.
\newblock \href {http://125.22.75.155:8080/handle/123456789/12198}
  {Ethnographic analysis and language with reference to {Malinowski}'s views}.
\newblock In Frank~R. Palmer, editor, \emph{Selected papers of {J}.{R}. {Firth}
  (1952-59)}, pages 137--168. Longman and Indiana University Press, London, UK,
  and Bloomington, IN, USA.

\bibitem[{Firth(1968{\natexlab{c}})}]{firth_linguistic_1968}
J.R. Firth. 1968{\natexlab{c}}.
\newblock \href {http://hdl.handle.net/123456789/12198} {Linguistic analysis as
  a study of meaning}.
\newblock In Frank~R. Palmer, editor, \emph{Selected papers of {J}.{R}. {Firth}
  (1952-59)}, pages 12--27. Longman and Indiana University Press, London, UK,
  and Bloomington, IN, USA.

\bibitem[{Firth(1968{\natexlab{d}})}]{firth_structural_1968}
J.R. Firth. 1968{\natexlab{d}}.
\newblock \href {http://hdl.handle.net/123456789/12198} {Structural
  linguistics}.
\newblock In Frank~R. Palmer, editor, \emph{Selected papers of {J}.{R}. {Firth}
  (1952-59)}, pages 35--53. Longman and Indiana University Press, London, UK,
  and Bloomington, IN, USA.

\bibitem[{Firth(1968{\natexlab{e}})}]{firth_treatment_1968}
J.R. Firth. 1968{\natexlab{e}}.
\newblock \href {http://hdl.handle.net/123456789/12198} {The treatment of
  language in general linguistics}.
\newblock In Frank~R. Palmer, editor, \emph{Selected papers of {J}.{R}. {Firth}
  (1952-59)}, pages 206--209. Longman and Indiana University Press, London, UK,
  and Bloomington, IN, USA.

\bibitem[{Garg et~al.(2018)Garg, Schiebinger, Jurafsky, and
  Zou}]{garg_word_2018}
Nikhil Garg, Londa Schiebinger, Dan Jurafsky, and James Zou. 2018.
\newblock \href {https://doi.org/10.1073/pnas.1720347115} {Word embeddings
  quantify 100 years of gender and ethnic stereotypes}.
\newblock \emph{Proceedings of the National Academy of Sciences of the United
  States of America}, 115(16):E3635--E3644.
\newblock ISBN: 9781720347118 publisher: National Academy of Sciences section:
  PNAS Plus.

\bibitem[{Garimella et~al.(2017)Garimella, Banea, and
  Mihalcea}]{garimella2017demographic}
Aparna Garimella, Carmen Banea, and Rada Mihalcea. 2017.
\newblock \href {https://aclanthology.org/D17-1242/} {Demographic-aware word
  associations}.
\newblock In \emph{Proceedings of the 2017 Conference on Empirical Methods in
  Natural Language Processing}, pages 2285--2295.

\bibitem[{Glenberg and Robertson(2000)}]{glenberg00}
Arthur~M. Glenberg and David~A. Robertson. 2000.
\newblock \href {https://doi.org/10.1006/jmla.2000.2714} {Symbol grounding and
  meaning: A comparison of high-dimensional and embodied theories of meaning}.
\newblock \emph{Journal of Memory and Language}, 43(3):379--401.

\bibitem[{Goldberg(2017)}]{goldberg_neural_2017}
Yoav Goldberg. 2017.
\newblock \href {https://doi.org/10.2200/S00762ED1V01Y201703HLT037} {Neural
  network models for natural language processing}.
\newblock \emph{Synthesis Lectures on Human Language Technologies},
  10(1):1--309.

\bibitem[{Goldsmith(2005)}]{goldsmith_legacy_2005}
John~A. Goldsmith. 2005.
\newblock \href {https://doi.org/10.1353/lan.2005.0127} {The legacy of {Zellig}
  {Harris}: {Language} and information into the 21st century, vol. 1:
  {Philosophy} of science, syntax and semantics (review)}.
\newblock \emph{Language}, 81(3):719--736.

\bibitem[{Gong et~al.(2020)Gong, Bhat, and Viswanath}]{gong_enriching_2020}
Hongyu Gong, Suma Bhat, and Pramod Viswanath. 2020.
\newblock \href {https://www.aclweb.org/anthology/2020.conll-1.1} {Enriching
  {Word} {Embeddings} with {Temporal} and {Spatial} {Information}}.
\newblock In \emph{Proceedings of the 24th {Conference} on {Computational}
  {Natural} {Language} {Learning}}, pages 1--11, Online. Association for
  Computational Linguistics.

\bibitem[{Goodfellow et~al.(2016)Goodfellow, Bengio, Courville, and
  Bengio}]{goodfellow2016deep}
Ian Goodfellow, Yoshua Bengio, Aaron Courville, and Yoshua Bengio. 2016.
\newblock \href {https://www.deeplearningbook.org/} {\emph{Deep learning}}.
\newblock MIT Press, Cambridge, MA, USA.

\bibitem[{Graves et~al.(2013)Graves, Mohamed, and Hinton}]{graves_speech_2013}
Alex Graves, Abdel-rahman Mohamed, and Geoffrey Hinton. 2013.
\newblock \href {https://doi.org/10.1109/ICASSP.2013.6638947} {Speech
  recognition with deep recurrent neural networks}.
\newblock In \emph{2013 {IEEE} {International} {Conference} on {Acoustics},
  {Speech} and {Signal} {Processing}}, pages 6645--6649.

\bibitem[{Grbovic and Cheng(2018)}]{grbovic_real-time_2018}
Mihajlo Grbovic and Haibin Cheng. 2018.
\newblock \href {https://doi.org/10.1145/3219819.3219885} {Real-time
  personalization using embeddings for search ranking at airbnb}.
\newblock In \emph{Proceedings of the 24th {ACM} {SIGKDD} {International}
  {Conference} on {Knowledge} {Discovery} \& {Data} {Mining} - {KDD} '18},
  pages 311--320, London, United Kingdom. ACM Press.

\bibitem[{Gumperz and Hymes(1972)}]{gumperz_directions_1972}
John~Joseph Gumperz and Dell Hymes, editors. 1972.
\newblock \href
  {https://www.worldcat.org/title/directions-in-sociolinguistics-the-ethnography-of-communication/oclc/281997}
  {\emph{Directions in sociolinguistics: {The} ethnography of communication}}.
\newblock Holt, Rinehart and Winston, New York, NY, USA.

\bibitem[{Habert and Zweigenbaum(2002)}]{nevin_contextual_2002}
Benoît Habert and Pierre Zweigenbaum. 2002.
\newblock \href {https://benjamins.com/catalog/cilt.228} {Contextual
  acquisition of information categories - {What} has been done and what can be
  done automatically?}
\newblock In Bruce~E. Nevin and Stephen~B. Johnson, editors, \emph{The {Legacy}
  of {Zellig} {Harris} - {Language} and information into the 21st century},
  volume 2: Mathematics and computability of language. John Benjamins
  Publishing Company.

\bibitem[{Halevy et~al.(2009)Halevy, Norvig, and
  Pereira}]{halevy_unreasonable_2009}
A.~Halevy, P.~Norvig, and F.~Pereira. 2009.
\newblock \href {https://doi.org/10.1109/MIS.2009.36} {The unreasonable
  effectiveness of data}.
\newblock \emph{IEEE Intelligent Systems}, 24(2):8--12.

\bibitem[{Halliday(1968)}]{halliday_users_1968}
M.A.K. Halliday. 1968.
\newblock \href
  {https://www.degruyter.com/document/doi/10.1515/9783110805376.139/html?lang=en}
  {The users and uses of language}.
\newblock In Joshua~A. Fishman, editor, \emph{Readings in the sociology of
  language}, pages 139--169. De Gruyter Mouton, New York, NY.

\bibitem[{Halliday(1991)}]{halliday_corpus_1991}
M.A.K. Halliday. 1991.
\newblock \href
  {https://www.taylorfrancis.com/chapters/edit/10.4324/9781315845890-12/corpus-studies-probabilistic-grammar-halliday}
  {Corpus studies and probabilistic grammar}.
\newblock In Karin Aijmer and Bengt Altenberg, editors, \emph{English {Corpus}
  {Linguistics}}, pages 30--43. Routledge, London, UK.

\bibitem[{Hamilton et~al.(2016)Hamilton, Leskovec, and
  Jurafsky}]{hamilton_diachronic_2016}
William~L. Hamilton, Jure Leskovec, and Dan Jurafsky. 2016.
\newblock \href {https://doi.org/10.18653/v1/P16-1141} {Diachronic word
  embeddings reveal statistical laws of semantic change}.
\newblock In \emph{Proceedings of the 54th {Annual} {Meeting} of the
  {Association} for {Computational} {Linguistics} ({Volume} 1: {Long}
  {Papers})}, pages 1489--1501, Berlin, Germany. Association for Computational
  Linguistics.

\bibitem[{Harris(1941)}]{harris_review_1941}
Zellig~S. Harris. 1941.
\newblock \href {https://doi.org/10.2307/409287} {Review of {Grundzüge} der
  {Phonologie}}.
\newblock \emph{Language}, 17(4):345--349.

\bibitem[{Harris(1951)}]{harris_methods_1951}
Zellig~S. Harris. 1951.
\newblock \href {https://psycnet.apa.org/record/1952-03375-000} {\emph{Methods
  in structural linguistics}}.
\newblock University of Chicago Press, Chicago, IL, USA.

\bibitem[{Harris(1954)}]{harris_distributional_1954}
Zellig~S. Harris. 1954.
\newblock \href {https://doi.org/10.1080/00437956.1954.11659520}
  {Distributional structure}.
\newblock \emph{WORD}, 10(2-3):146--162.

\bibitem[{Harris(1968)}]{harris_mathematical_1968}
Zellig~S. Harris. 1968.
\newblock \href
  {https://www.worldcat.org/title/mathematical-structures-of-language/oclc/00439265}
  {\emph{Mathematical structures of language}}.
\newblock John Wiley \& Sons, New York, NY, USA.

\bibitem[{Harris(1988)}]{harris_language_1988}
Zellig~S. Harris. 1988.
\newblock \href
  {https://books.google.com/books/about/Language_and_Information.html?id=2l2BQgAACAAJ}
  {\emph{Language and information}}.
\newblock Columbia University Press, New York, NY, USA.

\bibitem[{Harris(1991)}]{harris_theory_1991}
Zellig~S. Harris. 1991.
\newblock \href
  {https://global.oup.com/academic/product/a-theory-of-language-and-information-9780198242246?resultsPerPage=20&prevNumResPerPage=20&lang=es&cc=gb}
  {\emph{A theory of language and information: A mathematical approach}}.
\newblock Clarendon Press, Oxford, UK.

\bibitem[{Harris et~al.(1988)Harris, Gottfried, Ryckman, Daladier, and
  Mattick}]{harris_form_1988}
Zellig~S. Harris, Michael Gottfried, Thomas Ryckman, Anne Daladier, and Paul
  Mattick. 1988.
\newblock \href {https://link.springer.com/book/10.1007/978-94-009-2837-4}
  {\emph{The form of information in science: Analysis of an immunology
  sublanguage}}.
\newblock Kluwer Academic Publishers, Amsterdam.

\bibitem[{Henderson(2020)}]{henderson_unstoppable_2020}
James Henderson. 2020.
\newblock \href {https://doi.org/10.18653/v1/2020.acl-main.561} {The
  unstoppable rise of computational linguistics in deep learning}.
\newblock In \emph{Proceedings of the 58th {Annual} {Meeting} of the
  {Association} for {Computational} {Linguistics}}, pages 6294--6306, Online.
  Association for Computational Linguistics.

\bibitem[{Hindle(1990)}]{Hindle90}
Donald Hindle. 1990.
\newblock \href {https://dl.acm.org/doi/10.3115/981823.981857} {Noun
  classification from predicate-argument structures}.
\newblock In \emph{28th Annual meeting of the Association for Computational
  Linguistics}, pages 268--275, Pittsburgh, PA, USA. Association for
  Computational Linguistics.

\bibitem[{Hirschman et~al.(1975)Hirschman, Grishman, and
  Sager}]{hirschman1975grammatically}
Lynette Hirschman, Ralph Grishman, and Naomi Sager. 1975.
\newblock \href
  {https://www.sciencedirect.com/science/article/pii/0306457375900333}
  {Grammatically-based automatic word class formation}.
\newblock \emph{Information Processing \& Management}, 11(1-2):39--57.

\bibitem[{Hofmann(1999)}]{hofmann_probabilistic_1999}
Thomas Hofmann. 1999.
\newblock \href {https://doi.org/10.1145/312624.312649} {Probabilistic latent
  semantic indexing}.
\newblock In \emph{Proceedings of the 22nd annual international {ACM} {SIGIR}
  conference on {Research} and development in information retrieval}, {SIGIR}
  '99, pages 50--57, New York, NY, USA. Association for Computing Machinery.

\bibitem[{Hovy(2018)}]{hovy_social_2018}
Dirk Hovy. 2018.
\newblock \href {https://doi.org/10.18653/v1/W18-1106} {The social and the
  neural network: How to make natural language processing about people again}.
\newblock In \emph{Proceedings of the {Second} {Workshop} on {Computational}
  {Modeling} of {People}'s {Opinions}, {Personality}, and {Emotions} in
  {Social} {Media}}, pages 42--49, New Orleans, Louisiana, USA. Association for
  Computational Linguistics.

\bibitem[{Johnson(2002)}]{nevin_computability_2002}
Stephen~B. Johnson. 2002.
\newblock \href {https://benjamins.com/catalog/cilt.228} {The {Computability}
  of {Operator} {Grammar}}.
\newblock In Bruce~E. Nevin and Stephen~B. Johnson, editors, \emph{The {Legacy}
  of {Zellig} {Harris} - {Language} and information into the 21st century},
  volume 2: Mathematics and computability of language. John Benjamins
  Publishing Company.
\newblock Publication Title: cilt.228.

\bibitem[{Joos(1950)}]{joos_description_1950}
Martin Joos. 1950.
\newblock \href {https://doi.org/10.1121/1.1906674} {Description of language
  design}.
\newblock \emph{The Journal of the Acoustical Society of America},
  22(6):701--707.

\bibitem[{Jurafsky and Martin(2009)}]{jurafsky_speech_2009}
Dan Jurafsky and James~H. Martin. 2009.
\newblock \href {https://web.stanford.edu/~jurafsky/slp3/} {\emph{Speech and
  Language Processing: {An} introduction to natural language processing,
  computational linguistics, and speech recognition}}.
\newblock Prentice Hall, Hoboken, NJ, USA.

\bibitem[{Jurafsky and Martin(2021)}]{jurafsky_speech_2021}
Dan Jurafsky and James~H. Martin. 2021.
\newblock \href {https://web.stanford.edu/~jurafsky/slp3/} {\emph{Speech and
  language processing}}, 3rd (draft) edition.

\bibitem[{Kittredge and Lehrberger(1982)}]{kittredge_sublanguage_1982}
Richard Kittredge and John Lehrberger. 1982.
\newblock \href
  {https://www.degruyter.com/document/doi/10.1515/9783110844818/html?lang=en}
  {\emph{Sublanguage: {Studies} of language in restricted semantic domains}}.
\newblock de Gruyter, Berlin.

\bibitem[{Kozlowski et~al.(2019)Kozlowski, Taddy, and
  Evans}]{kozlowski_geometry_2019}
Austin~C. Kozlowski, Matt Taddy, and James~A. Evans. 2019.
\newblock \href {https://doi.org/10.1177/0003122419877135} {The geometry of
  culture: Analyzing the meanings of class through word embeddings}.
\newblock \emph{American Sociological Review}.
\newblock Publisher: SAGE PublicationsSage CA: Los Angeles, CA.

\bibitem[{Kutuzov et~al.(2018)Kutuzov, Øvrelid, Szymanski, and
  Velldal}]{kutuzov_diachronic_2018}
Andrey Kutuzov, Lilja Øvrelid, Terrence Szymanski, and Erik Velldal. 2018.
\newblock \href {https://www.aclweb.org/anthology/C18-1117} {Diachronic word
  embeddings and semantic shifts: a survey}.
\newblock In \emph{Proceedings of the 27th {International} {Conference} on
  {Computational} {Linguistics}}, pages 1384--1397, Santa Fe, New Mexico, USA.
  Association for Computational Linguistics.

\bibitem[{Labov(1966/2006)}]{labov_social_2006}
William Labov. 1966/2006.
\newblock \href
  {https://www.cambridge.org/core/books/social-stratification-of-english-in-new-york-city/F9CA8EE7DAA60345E838A9DA309DB8ED}
  {\emph{The social stratification of English in {New} {York} {City}}}, 2nd
  edition.
\newblock Cambridge University Press, Cambridge, UK.

\bibitem[{Labov(1972)}]{labov_sociolinguistic_1972}
William Labov. 1972.
\newblock \href
  {https://books.google.com/books/about/Sociolinguistic_Patterns.html?id=hD0PNMu8CfQC}
  {\emph{Sociolinguistic patterns}}.
\newblock University of Pennsylvania press, Philadelphia, PA, USA.

\bibitem[{Landauer and Dumais(1997)}]{landauer_solution_1997}
Thomas~K. Landauer and Susan~T. Dumais. 1997.
\newblock \href {https://doi.org/10.1037/0033-295X.104.2.211} {A solution to
  {Plato}'s problem: {The} latent semantic analysis theory of acquisition,
  induction, and representation of knowledge}.
\newblock \emph{Psychological Review}, 104(2):211--240.

\bibitem[{Le and Mikolov(2014)}]{le_distributed_2014}
Quoc Le and Tomas Mikolov. 2014.
\newblock \href {https://proceedings.mlr.press/v32/le14.html} {Distributed
  representations of sentences and documents}.
\newblock In \emph{Proceedings of the 31st {International} {Conference} on
  {Machine} {Learning}}, pages 1188--1196. PMLR.

\bibitem[{Levy and Goldberg(2014)}]{levy_dependency-based_2014}
Omer Levy and Yoav Goldberg. 2014.
\newblock \href {https://doi.org/10.3115/v1/P14-2050} {Dependency-based word
  embeddings}.
\newblock In \emph{Proceedings of the 52nd {Annual} {Meeting} of the
  {Association} for {Computational} {Linguistics} ({Volume} 2: {Short}
  {Papers})}, pages 302--308, Baltimore, MD, USA. Association for Computational
  Linguistics.

\bibitem[{Levy et~al.(2015)Levy, Goldberg, and Dagan}]{levy_improving_2015}
Omer Levy, Yoav Goldberg, and Ido Dagan. 2015.
\newblock \href {https://doi.org/10.1162/tacl_a_00134} {Improving
  distributional similarity with lessons learned from wor embeddings}.
\newblock \emph{Transactions of the Association for Computational Linguistics},
  3:211--225.

\bibitem[{Li et~al.(2016)Li, Galley, Brockett, Spithourakis, Gao, and
  Dolan}]{li2016persona}
Jiwei Li, Michel Galley, Chris Brockett, Georgios~P Spithourakis, Jianfeng Gao,
  and Bill Dolan. 2016.
\newblock \href {https://aclanthology.org/P16-1094/} {A persona-based neural
  conversation model}.
\newblock In \emph{Proceedings of the 54th Annual Meeting of the Association
  for Computational Linguistics}, pages 994--1003.

\bibitem[{Lu et~al.(2020)Lu, Mardziel, Wu, Amancharla, and
  Datta}]{lu_gender_2020}
Kaiji Lu, Piotr Mardziel, Fangjing Wu, Preetam Amancharla, and Anupam Datta.
  2020.
\newblock \href {https://doi.org/10.1007/978-3-030-62077-6_14} {Gender bias in
  neural natural language processing}.
\newblock In Vivek Nigam, Tajana Ban~Kirigin, Carolyn Talcott, Joshua Guttman,
  Stepan Kuznetsov, Boon Thau~Loo, and Mitsuhiro Okada, editors, \emph{Logic,
  {Language}, and {Security}: {Essays} {Dedicated} to {Andre} {Scedrov} on the
  {Occasion} of {His} 65th {Birthday}}, Lecture {Notes} in {Computer}
  {Science}, pages 189--202. Springer International Publishing, New York, NY,
  USA.

\bibitem[{Lucy and Bamman(2021)}]{lucy2021characterizing}
Li~Lucy and David Bamman. 2021.
\newblock \href
  {https://direct.mit.edu/tacl/article/doi/10.1162/tacl_a_00383/101877/Characterizing-English-Variation-across-Social}
  {Characterizing english variation across social media communities with bert}.
\newblock \emph{Transactions of the Association for Computational Linguistics},
  9:538--556.

\bibitem[{Léon(2007)}]{leon_linguistic_2007}
Jacqueline Léon. 2007.
\newblock \href {https://halshs.archives-ouvertes.fr/halshs-00220455} {From
  linguistic events and restricted languages to registers: {Firthian} legacy
  and corpus linguistics}.
\newblock \emph{The Henry Sweet Society Bulletin}, 49(1):5--25.

\bibitem[{Léon(2008)}]{leon_empirical_2008}
Jacqueline Léon. 2008.
\newblock \href {https://halshs.archives-ouvertes.fr/halshs-00370872}
  {Empirical {Traditions} of {Computer}-{Based} {Methods}. {Firth}'s
  {Restricted} {Languages} and {Harris}' {Sublanguages}}.
\newblock \emph{Beiträge zur Geschichte der Sprachwissenschaft}, 18(2):259.

\bibitem[{Léon(2011)}]{leon_z_2011}
Jacqueline Léon. 2011.
\newblock \href
  {https://www.jbe-platform.com/content/books/9789027287175-sihols.115.39leo}
  {Z. {S}. {Harris} and the semantic turn of mathematical information theory}.
\newblock \emph{History of Linguistics 2008}, pages 449--458.
\newblock Publisher: John Benjamins.

\bibitem[{Léon(2021)}]{leon_automating_2021}
Jacqueline Léon. 2021.
\newblock \href {https://link.springer.com/book/10.1007/978-3-030-70642-5}
  {\emph{Automating linguistics}}.
\newblock Springer Nature, London, UK.

\bibitem[{Manning(2015)}]{manning_computational_2015}
Christopher~D. Manning. 2015.
\newblock \href {https://doi.org/10.1162/COLI_a_00239} {Computational
  linguistics and deep learning}.
\newblock \emph{Computational Linguistics}, 41(4):701--707.

\bibitem[{Manning and Schütze(1999)}]{manning_foundations_1999}
Christopher~D. Manning and Hinrich Schütze. 1999.
\newblock \href
  {https://books.google.com/books?hl=en&lr=&id=3qnuDwAAQBAJ&oi=fnd&pg=PT12&dq=Foundations+of+statistical+natural+language+processing&ots=ysKYp6GqRW&sig=ceOjodpD4evCKnJmt1JbtAA-C5M#v=onepage&q=Foundations\%20of\%20statistical\%20natural\%20language\%20processing&f=false}
  {\emph{Foundations of statistical natural language processing}}.
\newblock MIT Press, Cambridge, MA, USA.

\bibitem[{Martinc et~al.(2020)Martinc, Kralj~Novak, and
  Pollak}]{martinc_leveraging_2020}
Matej Martinc, Petra Kralj~Novak, and Senja Pollak. 2020.
\newblock \href {https://aclanthology.org/2020.lrec-1.592} {Leveraging
  contextual embeddings for detecting diachronic semantic shift}.
\newblock In \emph{Proceedings of the 12th {Language} {Resources} and
  {Evaluation} {Conference}}, pages 4811--4819, Marseille, France. European
  Language Resources Association.

\bibitem[{McKenzie and Adams(2021)}]{mckenzie_natural_2021}
Grant McKenzie and Benjamin Adams. 2021.
\newblock \href
  {https://gistbok.ucgis.org/bok-topics/natural-language-processing-giscience-applications}
  {Natural language processing in {GIScience} applications}.
\newblock In John~P. Wilson, editor, \emph{Geographic {Information} {Science}
  \& {Technology} {Body} of {Knowledge}}, 4th quarter 2021 edition edition.

\bibitem[{Mendelsohn et~al.(2020)Mendelsohn, Tsvetkov, and
  Jurafsky}]{mendelsohn2020framework}
Julia Mendelsohn, Yulia Tsvetkov, and Dan Jurafsky. 2020.
\newblock \href
  {https://www.frontiersin.org/articles/10.3389/frai.2020.00055/full} {A
  framework for the computational linguistic analysis of dehumanization}.
\newblock \emph{Frontiers in Artificial Intelligence}, 3:1--24.

\bibitem[{Mikolov et~al.(2013)Mikolov, Sutskever, Chen, Corrado, and
  Dean}]{mikolov_distributed_2013}
Tomas Mikolov, Ilya Sutskever, Kai Chen, Greg~S Corrado, and Jeff Dean. 2013.
\newblock \href
  {https://papers.nips.cc/paper/2013/hash/9aa42b31882ec039965f3c4923ce901b-Abstract.html}
  {Distributed representations of words and phrases and their
  compositionality}.
\newblock In \emph{Advances in {Neural} {Information} {Processing} {Systems}},
  volume~26. Curran Associates, Inc., Redhook, NY, USA.

\bibitem[{Mitchell(1957)}]{mitchell_language_1957}
T.~F Mitchell. 1957.
\newblock \href
  {https://www.worldcat.org/title/language-of-buying-and-selling-in-cyrenaica-a-situational-statement/oclc/873196683}
  {The language of buying and selling in {Cyrenaica}: A situational statement}.
\newblock \emph{Hespéris - Archives Berbères et Bulletin d'Institut des
  Hautes Études Marocaines}, 44:31--71.

\bibitem[{Nevin(2010)}]{nevin2010}
Bruce Nevin. 2010.
\newblock \href
  {https://library.oapen.org/bitstream/handle/20.500.12657/44084/external_content.pdf?sequence=1#page=116}
  {Noam and {Z}ellig}.
\newblock In Douglas~A. Kibbee, editor, \emph{Chomskyan (R)evolutions}, pages
  103--168. John Benjamins Philadelphia.

\bibitem[{Nevin(1993)}]{nevin_minimalist_1993}
Bruce~E. Nevin. 1993.
\newblock \href {https://doi.org/10.1075/hl.20.2-3.06nev} {A minimalist program
  for linguistics: {The} work of {Zellig} {Harris} on meaning and information}.
\newblock \emph{Historiographia Linguistica}, 20(2-3):355--398.

\bibitem[{Nguyen et~al.(2021)Nguyen, Rosseel, and
  Grieve}]{nguyen_learning_2021}
Dong Nguyen, Laura Rosseel, and Jack Grieve. 2021.
\newblock \href {https://doi.org/10.18653/v1/2021.naacl-main.50} {On learning
  and representing social meaning in {NLP}: a sociolinguistic perspective}.
\newblock In \emph{Proceedings of the 2021 {Conference} of the {North}
  {American} {Chapter} of the {Association} for {Computational} {Linguistics}:
  {Human} {Language} {Technologies}}, pages 603--612, Online. Association for
  Computational Linguistics.

\bibitem[{Norvig(2012)}]{norvig_colorless_2012}
Peter Norvig. 2012.
\newblock \href
  {https://doi.org/https://doi.org/10.1111/j.1740-9713.2012.00590.x} {Colorless
  green ideas learn furiously: {Chomsky} and the two cultures of statistical
  learning}.
\newblock \emph{Significance}, 9(4):30--33.

\bibitem[{Oyelaran(1967)}]{oyelaran_aspects_1967}
Olasope~O. Oyelaran. 1967.
\newblock \href {https://doi.org/10.1080/00437956.1967.11435497} {Aspects of
  linguistic theory in {Firthian} linguistics}.
\newblock \emph{\textit{WORD}}, 23(1-3):428--452.

\bibitem[{Palmer(1968)}]{firth_introduction_1968}
F.R. Palmer. 1968.
\newblock \href {https://books.google.fi/books?id=Uxu3AAAAIAAJ} {Introduction}.
\newblock In J.R. Firth, editor, \emph{Selected {Papers} of {J}.{R}. {Firth},
  1952-59}. Longmans, Green and Co., London, UK.

\bibitem[{Pandit(1970)}]{pandit_review_1970}
P.~B. Pandit. 1970.
\newblock \href {http://www.jstor.org/stable/4175086} {Review of {Selected}
  {Papers} of {J}. {R}. {Firth} 1952-59}.
\newblock \emph{Journal of Linguistics}, 6(2):280--284.
\newblock Publisher: Cambridge University Press.

\bibitem[{Peters et~al.(2018)Peters, Neumann, Iyyer, Gardner, Clark, Lee, and
  Zettlemoyer}]{peters_deep_2018}
Matthew Peters, Mark Neumann, Mohit Iyyer, Matt Gardner, Christopher Clark,
  Kenton Lee, and Luke Zettlemoyer. 2018.
\newblock \href {https://doi.org/10.18653/v1/N18-1202} {Deep contextualized
  word representations}.
\newblock In \emph{Proceedings of the 2018 {Conference} of the {North}
  {American} {Chapter} of the {Association} for {Computational} {Linguistics}:
  {Human} {Language} {Technologies}, {Volume} 1 ({Long} {Papers})}, pages
  2227--2237, New Orleans, LA, USA. Association for Computational Linguistics.

\bibitem[{Plug(2008)}]{plug_j_2008}
Leendert Plug. 2008.
\newblock \href {https://doi.org/10.1111/j.1467-968X.2008.00203.x} {J. {R}.
  {Firth}: a new biography}.
\newblock \emph{Transactions of the Philological Society}, 106(3):337--374.

\bibitem[{Raina et~al.(2007)Raina, Battle, Lee, Packer, and
  Ng}]{raina_self-taught_2007}
Rajat Raina, Alexis Battle, Honglak Lee, Benjamin Packer, and Andrew~Y. Ng.
  2007.
\newblock \href {https://doi.org/10.1145/1273496.1273592} {Self-taught
  learning: Transfer learning from unlabeled data}.
\newblock In \emph{Proceedings of the 24th international conference on
  {Machine} learning}, {ICML} '07, pages 759--766, New York, NY, USA.
  Association for Computing Machinery.

\bibitem[{Ramesh et~al.(2022)Ramesh, Dhariwal, Nichol, Chu, and
  Chen}]{ramesh_hierarchical_2022}
Aditya Ramesh, Prafulla Dhariwal, Alex Nichol, Casey Chu, and Mark Chen. 2022.
\newblock \href {https://cdn.openai.com/papers/dall-e-2.pdf} {Hierarchical
  text-conditional image generation with {CLIP} latents}.
\newblock \emph{OpenAI}, pages 1--26.

\bibitem[{Robins(1953)}]{robins_phonology_1953}
R.~H. Robins. 1953.
\newblock \href {http://www.jstor.org/stable/608889} {The phonology of the
  nasalized verbal forms in sundanese}.
\newblock \emph{Bulletin of the School of Oriental and African Studies,
  University of London}, 15(1):138--145.

\bibitem[{Robins(1971)}]{ardener_malinowski_1971}
R.~H. Robins. 1971.
\newblock \href
  {https://www.taylorfrancis.com/chapters/mono/10.4324/9781315017617-11/malinowski-firth-context-situation-edwin-ardener}
  {Malinowski, {Firth}, and the ``{context} of {situation}''}.
\newblock In Edwin Ardener, editor, \emph{Social {Anthropology} and
  {Language}}. Routledge, London, UK.

\bibitem[{Robins(1997)}]{robins_contribution_1997}
R.~H. Robins. 1997.
\newblock \href {https://doi.org/10.1016/S0024-3841(96)00042-3} {The
  contribution of {John} {Rupert} {Firth} to linguistics in the first fifty
  years of \textit{Lingua}}.
\newblock \emph{Lingua}, 100(1):205--222.

\bibitem[{Rubenstein and Goodenough(1965)}]{rubenstein_contextual_1965}
Herbert Rubenstein and John~B. Goodenough. 1965.
\newblock \href {https://doi.org/10.1145/365628.365657} {Contextual correlates
  of synonymy}.
\newblock \emph{Communications of the ACM}, 8(10):627--633.

\bibitem[{Russell and Norvig(2020)}]{russell_artificial_2020}
Stuart~J. Russell and Peter Norvig. 2020.
\newblock \href {http://aima.cs.berkeley.edu/} {\emph{Artificial intelligence:
  {A} modern approach}}, 4th edition.
\newblock Pearson, Hoboken, NJ, USA.

\bibitem[{Sager(1975)}]{sager_sublanguage_1975}
Naomi Sager. 1975.
\newblock \href
  {https://asistdl.onlinelibrary.wiley.com/doi/abs/10.1002/asi.4630260104?casa_token=kvdpTHK_zLQAAAAA:-IUgIdQjDjXPVRToOioLAZUW5QbmxYR75JcHN7LCSGTob6tYPWNMZHGThaP6mW0eQwMO0W27w45QGLc}
  {Sublanguage grammers in science information processing}.
\newblock \emph{Journal of the American Society for Information Science},
  26(1):10--16.

\bibitem[{Sager(1981)}]{sager_natural_1981}
Naomi Sager. 1981.
\newblock \href
  {https://www.abebooks.com/9780201067699/Natural-Language-Information-Processing-Computer-0201067692/plp}
  {\emph{Natural language information processing: A computer grammar of
  {English} and its applications}}.
\newblock Addison-Wesley Publishing Company, Advanced Book Program, Boston, MA,
  USA.

\bibitem[{Sahlgren(2008)}]{sahlgren_distributional_2008}
Magnus Sahlgren. 2008.
\newblock \href
  {https://www.diva-portal.org/smash/get/diva2:1041938/FULLTEXT01.pdf} {The
  distributional hypothesis}.
\newblock \emph{Rivista di Linguistica}, (20.1):33--53.

\bibitem[{Sap et~al.(2020)Sap, Gabriel, Qin, Jurafsky, Smith, and
  Choi}]{sap_social_2020}
Maarten Sap, Saadia Gabriel, Lianhui Qin, Dan Jurafsky, Noah~A. Smith, and
  Yejin Choi. 2020.
\newblock \href {https://doi.org/10.18653/v1/2020.acl-main.486} {Social bias
  frames: Reasoning about social and power implcations of language}.
\newblock In \emph{Proceedings of the 58th {Annual} {Meeting} of the
  {Association} for {Computational} {Linguistics}}, pages 5477--5490, Online.
  Association for Computational Linguistics.

\bibitem[{Sch{\"u}tze and Pedersen(1993)}]{schutze_vector_1993}
Hinrich Sch{\"u}tze and Jan Pedersen. 1993.
\newblock \href
  {https://citeseerx.ist.psu.edu/viewdoc/download?doi=10.1.1.54.6279&rep=rep1&type=pdf}
  {A vector model for syntagmatic and paradigmatic relatedness}.
\newblock In \emph{Proceedings of the 9th Annual Conference of the UW Centre
  for the New OED and Text Research}, pages 104--113, Oxford, UK.

\bibitem[{Schütze(1992)}]{schutze_dimensions_1992}
Hinrich Schütze. 1992.
\newblock \href {https://doi.org/10.1109/SUPERC.1992.236684} {Dimensions of
  meaning}.
\newblock In \emph{Supercomputing '92: Proceedings of the 1992 ACM/IEEE
  Conference on Supercomputing}, pages 787--796. IEEE Computer Society,
  Washington, DC, USA.

\bibitem[{Schütze(1993)}]{schutze_word_1993}
Hinrich Schütze. 1993.
\newblock \href {https://dl.acm.org/doi/10.5555/645753.758140} {Word space}.
\newblock In \emph{Advances in {Neural} {Information} {Processing} {Systems}},
  volume~5, pages 895--902. Morgan-Kaufmann, San Francisco, CA, USA.

\bibitem[{Senis(2015)}]{senis_contribution_2015}
Angela Senis. 2015.
\newblock \href {https://hal.archives-ouvertes.fr/hal-01515010} {The
  contribution of {John} {Rupert} {Firth} to the history of linguistics and the
  rejection of the phoneme theory}.
\newblock In \emph{{ConSOLE} {XXIII} (23rd {Conference} of the {Student}
  {Organization} of {Linguistics} in {Europe})}, pages 273--293, Paris, France.
  Leiden University Centre for Linguistics.

\bibitem[{Shannon(1948)}]{shannon_mathematical_1948}
C.~E. Shannon. 1948.
\newblock \href {https://doi.org/10.1002/j.1538-7305.1948.tb00917.x} {A
  mathematical theory of communication}.
\newblock \emph{The Bell System Technical Journal}, 27(4):623--656.

\bibitem[{Shannon(1945)}]{shannon_mathematical_1945}
C.E. Shannon. 1945.
\newblock \href {https://evervault.com/papers/shannon} {A mathematical theory
  of cryptography - {C}ase 10878}.
\newblock Technical report, Bell Telephone Laboratories, Princeton Libraries.

\bibitem[{Silverstein(2003)}]{silverstein_indexical_2003}
Michael Silverstein. 2003.
\newblock \href {https://doi.org/10.1016/S0271-5309(03)00013-2} {Indexical
  order and the dialectics of sociolinguistic life}.
\newblock \emph{Language \& Communication}, 23(3-4):193--229.

\bibitem[{Spinde et~al.(2021)Spinde, Rudnitckaia, Hamborg, and
  Gipp}]{spinde2021identification}
Timo Spinde, Lada Rudnitckaia, Felix Hamborg, and Bela Gipp. 2021.
\newblock \href
  {https://link.springer.com/chapter/10.1007/978-3-030-71305-8_17}
  {Identification of biased terms in news articles by comparison of
  outlet-specific word embeddings}.
\newblock In \emph{International Conference on Information}, pages 215--224.
  Springer.

\bibitem[{Tamari et~al.(2020)Tamari, Shani, Hope, Petruck, Abend, and
  Shahaf}]{tamari_language_2020}
Ronen Tamari, Chen Shani, Tom Hope, Miriam R~L Petruck, Omri Abend, and Dafna
  Shahaf. 2020.
\newblock \href {https://doi.org/10.18653/v1/2020.acl-main.559} {Language
  (re)modelling: {Towards} embodied language understanding}.
\newblock In \emph{Proceedings of the 58th {Annual} {Meeting} of the
  {Association} for {Computational} {Linguistics}}, pages 6268--6281, Online.
  Association for Computational Linguistics.

\bibitem[{Tenney et~al.(2019)Tenney, Xia, Chen, Wang, Poliak, McCoy, Kim,
  Van~Durme, Bowman, Das, and Pavlick}]{tenney-EtAl:2019:ICLR}
Ian Tenney, Patrick Xia, Berlin Chen, Alex Wang, Adam Poliak, R~Thomas McCoy,
  Najoung Kim, Benjamin Van~Durme, Samuel~R Bowman, Dipanjan Das, and Ellie
  Pavlick. 2019.
\newblock \href {https://arxiv.org/abs/1905.06316} {What do you learn from
  context? {Probing} for sentence structure in contextualized word
  representations}.
\newblock In \emph{Proceedings of the 2019 International Conference on Learning
  Representations}, pages 1--17, New Orleans, LA, USA.

\bibitem[{Thomas(2011)}]{thomas_fifty_2011}
Margaret Thomas. 2011.
\newblock \href {https://doi.org/10.4324/9780203814482} {\emph{Fifty key
  thinkers on language and linguistics}}.
\newblock Routledge, London, UK.

\bibitem[{Trott et~al.(2020)Trott, Torrent, Chang, and
  Schneider}]{trott_reconstruing_2020}
Sean Trott, Tiago~Timponi Torrent, Nancy Chang, and Nathan Schneider. 2020.
\newblock \href {https://doi.org/10.18653/v1/2020.acl-main.462}
  {({Re})construing meaning in {NLP}}.
\newblock In \emph{Proceedings of the 58th {Annual} {Meeting} of the
  {Association} for {Computational} {Linguistics}}, pages 5170--5184, Online.
  Association for Computational Linguistics.

\bibitem[{Vaswani et~al.(2017)Vaswani, Shazeer, Parmar, Uszkoreit, Jones,
  Gomez, Kaiser, and Polosukhin}]{vaswani_attention_2017}
Ashish Vaswani, Noam Shazeer, Niki Parmar, Jakob Uszkoreit, Llion Jones,
  Aidan~N. Gomez, Łukasz Kaiser, and Illia Polosukhin. 2017.
\newblock \href
  {https://papers.nips.cc/paper/2017/hash/3f5ee243547dee91fbd053c1c4a845aa-Abstract.html}
  {Attention is all you need}.
\newblock \emph{Advances in Neural Information Processing Systems},
  30:6000--6010.

\bibitem[{Weaver(1952)}]{weaver_translation_1952}
Warren Weaver. 1952.
\newblock \href {https://aclanthology.org/1952.earlymt-1.1} {Translation}.
\newblock In \emph{Proceedings of the {Conference} on {Mechanical}
  {Translation}}, pages 1--12, Cambridge, MA, USA.

\bibitem[{Whitehead(1938)}]{whitehead_modes_1938}
Alfred~North Whitehead. 1938.
\newblock \href
  {https://www.simonandschuster.com/books/Modes-of-Thought/Alfred-North-Whitehead/9780029352106}
  {\emph{Modes of thought}}.
\newblock Cambridge University Press, Cambridge, UK.

\bibitem[{Whitehead(1957)}]{whitehead_process_1957}
Alfred~North Whitehead. 1957.
\newblock \emph{Process and {Reality}}.
\newblock Macmillan, New York, NY.

\bibitem[{Wittgenstein(1953)}]{wittgenstein_philosophical_1953}
Ludwig Wittgenstein. 1953.
\newblock \href
  {https://www.worldcat.org/title/philosophical-investigations/oclc/371912}
  {\emph{Philosophical investigations}}.
\newblock Macmillan, New York, NY, USA.

\bibitem[{Yan et~al.(2017)Yan, Janowicz, Mai, and Gao}]{yan_itdl_2017}
Bo~Yan, Krzysztof Janowicz, Gengchen Mai, and Song Gao. 2017.
\newblock \href {https://doi.org/10.1145/3139958.3140054} {From {ITDL} to
  {Place2Vec}: {Reasoning} about place type similarity and relatedness by
  learning embeddings from augmenteed spatial contexts}.
\newblock In \emph{Proceedings of the 25th {ACM} {SIGSPATIAL} {International}
  {Conference} on {Advances} in {Geographic} {Information} {Systems} -
  {SIGSPATIAL}'17}, pages 1--10, Redondo Beach, CA, USA. ACM Press.

\bibitem[{Yao et~al.(2020)Yao, Dou, and Wen}]{yao2020employing}
Jing Yao, Zhicheng Dou, and Ji-Rong Wen. 2020.
\newblock \href {https://dl.acm.org/doi/abs/10.1145/3397271.3401153} {Employing
  personal word embeddings for personalized search}.
\newblock In \emph{Proceedings of the 43rd International ACM SIGIR Conference
  on Research and Development in Information Retrieval}, pages 1359--1368.

\end{thebibliography}
\bibliographystyle{acl_natbib}

\end{document}